\documentclass[10pt,journal,compsoc]{IEEEtran}

\usepackage[ruled,vlined,linesnumbered]{algorithm2e}
\usepackage{listings}
\usepackage{xcolor}
\usepackage{amssymb}
\usepackage{pifont}
\newcommand{\cmark}{\ding{51}}%
\newcommand{\xmark}{\ding{53}}%
\usepackage{multirow}
\usepackage{dblfloatfix}  
\usepackage{hyperref}

\definecolor{codegreen}{rgb}{0,0.6,0}
\definecolor{codegray}{rgb}{0.5,0.5,0.5}
\definecolor{codepurple}{rgb}{0.58,0,0.82}
\definecolor{backcolour}{rgb}{0.95,0.95,0.92}

\lstdefinestyle{mystyle}{
    backgroundcolor=\color{backcolour},   
    commentstyle=\color{codegreen},
    keywordstyle=\color{magenta},
    numberstyle=\tiny\color{codegray},
    stringstyle=\color{codepurple},
    basicstyle=\ttfamily\footnotesize,
    breakatwhitespace=false,         
    breaklines=true,                 
    captionpos=b,                    
    keepspaces=true,                 
    numbers=left,                    
    numbersep=5pt,                  
    showspaces=false,                
    showstringspaces=false,
    showtabs=false,                  
    tabsize=2
}

\usepackage{makecell}
\usepackage{caption}
\usepackage{floatrow} 
\usepackage{siunitx}
\usepackage{wrapfig}
\usepackage{multicol}
\usepackage{booktabs}
\usepackage{subcaption}
\usepackage{amsmath}
\usepackage{amssymb}
\usepackage{cleveref}

\ifCLASSOPTIONcompsoc
  \usepackage[nocompress]{cite}
\else
  \usepackage{cite}
\fi

\ifCLASSINFOpdf
  \usepackage[pdftex]{graphicx}

\else

\fi

\begin{document}

\title{From Keypoints to Object Landmarks via Self-Training Correspondence: A novel approach to Unsupervised Landmark Discovery}

\author{Dimitrios Mallis, Enrique Sanchez, Matt Bell and Georgios Tzimiropoulos
\IEEEcompsocitemizethanks{
\IEEEcompsocthanksitem Dimitrios Mallis is with the Computer Vision Lab, University of Nottingham, NG8 1BB, UK.\protect\\
E-mail: malldimi1@gmail.com
\IEEEcompsocthanksitem Enrique Sanchez is with
Samsung AI Center Cambridge, CB1 2RE, UK.\protect\\
E-mail: kike.sanc@gmail.com
\IEEEcompsocthanksitem Matt Bell is with the Department of Animal and Agriculture, Hartpury University, GL19 3BE , UK.\protect\\
E-mail: matt.bell@hartpury.ac.uk
\IEEEcompsocthanksitem Georgios Tzimiropoulos is with Samsung AI Center Cambridge, Cambridge CB1 2RE, U.K., and also with the School of Electronic Engineering and
Computer Science, Queen Mary University of London, E1 4NS, U.K.\protect\\
E-mail: g.tzimiropoulos@qmul.ac.uk
}
\thanks{Manuscript accepted on January 2023.}
}

\markboth{IEEE Transactions on Pattern Analysis and Machine Intelligence}%
{Shell \MakeLowercase{\textit{et al.}}: Bare Advanced Demo of IEEEtran.cls for IEEE Computer Society Journals}

\IEEEtitleabstractindextext{%

\begin{abstract}
This paper proposes a novel paradigm for the unsupervised learning of object landmark detectors. Contrary to existing methods that build on auxiliary tasks such as image generation or equivariance, we propose a self-training approach where, departing from generic keypoints, a landmark detector and descriptor is trained to improve itself, tuning the keypoints into distinctive landmarks. To this end, we propose an iterative algorithm that alternates between producing new pseudo-labels through feature clustering and learning distinctive features for each pseudo-class through contrastive learning. With a shared backbone for the landmark detector and descriptor, the keypoint locations progressively converge to stable landmarks, filtering those less stable. Compared to previous works, our approach can learn points that are more flexible in terms of capturing large viewpoint changes. We validate our method on a variety of difficult datasets, including LS3D, BBCPose, Human3.6M and PennAction, achieving new state of the art results. Code and models can be found at \url{https://github.com/dimitrismallis/KeypointsToLandmarks}.
\end{abstract}

\begin{IEEEkeywords}
Unsupervised Landmark Discovery, Self-Training, Clustering, Correspondence, Keypoints
\end{IEEEkeywords}}

\maketitle

\IEEEdisplaynontitleabstractindextext

\IEEEpeerreviewmaketitle

\ifCLASSOPTIONcompsoc
\IEEEraisesectionheading{\section{Introduction}\label{sec:introduction}}
\else
\section{Introduction}
\label{sec:introduction}
\fi

\IEEEPARstart{O}{bject} parts, also known as landmarks, convey information about the shape and spatial configuration of an object in 3D space, especially for deformable objects like the human face, body and hand. Landmarks represent the locations of the specific parts with particular semantic meaning and thus follow an indexed configuration that is often manually designed.

The goal of landmark detection is to have a model that, for a particular instance of an object can estimate the locations of its parts or landmarks. Research in this field is mainly driven by supervised approaches, where sufficient amount of human-annotated data is provided. Common object categories used in part-based detection are faces~\cite{Bulat2018SuperFANIF,Dong_2018_CVPR, tzimiropoulos2017fast, sanchez2017functional} or human bodies~\cite{ Newell2016StackedHN, xiao2018simple}, where thousands of annotated images with landmarks are available. However, as in many other Computer Vision disciplines, relying on human annotations to develop novel detectors is costly, and hence alternative methods based on unsupervised learning are being explored.

\begin{figure}[!t]
\centering
\includegraphics[width=0.9\linewidth]{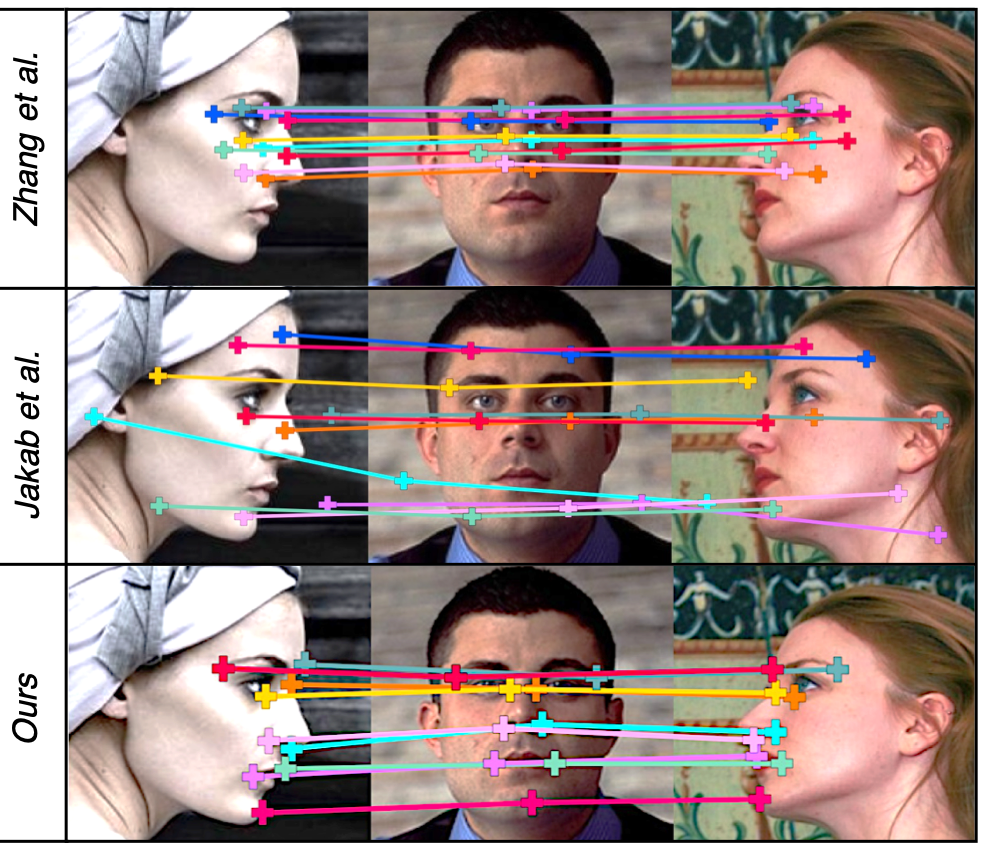}
\caption{Contrary to previous works that fail to cope with large viewpoint changes~\cite{jakab2018unsupervised} or that fail to deal with object symmetries~\cite{Zhang2018UnsupervisedDO}, our method finds correspondence across large viewpoint changes, leading to the discovery of landmarks that better represent the object's geometry.}

\label{fig:correspondancefigure}
\end{figure}

Unsupervised learning of object landmarks from a first glance seems an impossible task. A human annotator has understanding of the notion of objects and their parts, viewpoint invariance, occlusion and self-occlusion as well as examples of which landmarks to annotate in their disposal. On the contrary, unsupervised learning often relies on an auxiliary or proxy task, whereby the target task naturally arises as a latent process. Some techniques are either based on learning strong representations that can be mapped to manual landmarks using few images~\cite{Thewlis2017UnsupervisedLOobjectframes} or on discovering the landmarks from raw images through auxiliary proxy losses, such as equivariance \cite{Thewlis2017UnsupervisedLO, Thewlis2017UnsupervisedLOobjectframes,Thewlis2019UnsupervisedLO }, or tasks such as image generation\cite{Zhang2018UnsupervisedDO,jakab2018unsupervised,NIPS2019_9505}. Methods based on the \textit{principle of equivariance} observe that a detector must be consistent under known synthetic image deformations and attempt to optimise such objective. Methods based on image generation rely on reconstructing a deformed image through a generator that is conditioned on the detector's output; the detector and generator communicate through a bottleneck aimed to \textit{distill the object's geometry}. For the generator to recover the input image from a deformed version of itself, the detector needs to produce meaningful landmarks.

While these approaches have shown good performance in limited scenarios where objects showcase little rigid deformation (frontal faces or bodies, shoes, cat faces, etc), they are limited, by definition, in two critical aspects. First, a proxy task does not enforce the explicit learning of object landmarks, and thus are prone to generate landmarks that would unlikely be selected by a human annotator. Second, these methods require synthetically generated deformations since local correspondences for unpaired images are not known in the unsupervised case. Learning from pairs of images where one is a synthetic deformation of the other results in representations with limited robustness to intraclass variation that may not generalise well to highly articulated objects like the human body, complicated backgrounds or large viewpoint changes (i.e. 3D rotations).

In this paper, we observe that, while landmark detectors are difficult to train in an unsupervised manner, generic keypoint detectors, on the contrary, are much simpler to obtain and thus propose a novel method that can convert the latter into the former. Generic keypoints, often also referred to as salient or interest points, are simply points in an image representing the locations where ``something occurs", i.e. where there is a variation on the appearance, an edge, etc. Beyond representing a geometric position in an image, keypoints are represented by a feature descriptor, which is often used to find correspondences across different images (e.g. parts of two different images corresponding to different views of a building). Generic keypoints can be directly computed using Sobel filters (e.g. SIFT) or by training a detector on synthetic image deformations and homographic recovery (e.g. SuperPoint~\cite{DeTone2018SuperPointSI}). 

Based on the similarities and differences between keypoints and landmarks, our goal in this paper is to convert a series of keypoints automatically detected for a given object category into semantically coherent landmarks that describe the object parts, filtering and refining during the training process the corresponding landmark locations. To this end, we propose a novel approach that a) discovers landmarks through self-training instead of auxiliary objectives and b) captures intraclass variation from random image pairs. 

Our main starting point consists of populating a dataset of images belonging to a target object category (e.g. faces, birds) with a set of keypoints. It is expected that some of these points will show consistency and will systematically overlap with what we would refer to as landmarks. From this initial setup, our goal is to develop a self-training approach that can be used to learn a landmark detector in a fully unsupervised manner. In particular, we introduce a network akin to that of SuperPoint~\cite{DeTone2018SuperPointSI} (i.e. with a detector head and a descriptor head) that learns iteratively, through self-training, to locate a set of keypoints and to assign to each a distinctive descriptor that is landmark-consistent. Our goal is then to turn a keypoint detector into a landmark detector where the points capture the semantic meaning of a particular object in an unsupervised manner and re-label the training data accordingly. Then, a simple landmark detector based on heatmap regression can be trained as the final network. To this end, we propose to iteratively alternate between \textbf{pseudo-labelling of keypoints along with correspondence recovery, through descriptor clustering}, and \textbf{model self-training with produced pseudo-labels}.

We observe that, compared to previous works, our proposed approach is capable of learning landmarks that are more flexible in terms of capturing changes in 3D viewpoint. See for example Fig.~\ref{fig:correspondancefigure}. We demonstrate some of the favourable properties of our method on a variety of difficult datasets including LS3D~\cite{Bulat2017HowFA}, BBCPose~\cite{Charles2013DomainAF}, Human3.6M~\cite{Ionescu2014Human36MLS} and PennAction~\cite{Zhang2013FromAT}, notably without utilizing temporal information.

This manuscript extends and modifies our prior work ~\cite{unsupervLandm2020} both methodologically and experimentally. In particular, while in ~\cite{unsupervLandm2020} the number of landmarks to be discovered was part of the algorithm, we opt for keeping them fixed as in prior work~\cite{Thewlis2017UnsupervisedLO,Zhang_2018_CVPR,jakab2018unsupervised,NIPS2019_9505} by using a two-way K-means clustering algorithm (Sec.~\ref{ssec:learning_detector}). In addition, we observe that the negative pair selection in~\cite{unsupervLandm2020} might lead to the sampling of negative pairs that only differ in their cluster assignment because they encode different viewpoints of the same landmark. To avoid such an effect, we modify the negative pair selection to account only for samples that come from the same image, ensuring negative pairs refer not only to different clusters, but also to different landmarks. Finally, rather than originally populating the descriptors with those of the keypoint detector, we opt for a warm-up strategy that removes the dependency of our method in the quality of the initial descriptors. Experimentally, we conduct a thorough ablation study and include results in the challenging human pose dataset PennAction~\cite{Zhang2013FromAT} as well as CatFaces~\cite{Zhang2008CatHD} and Caltech-UCSD Birds \cite{WelinderEtal2010}. The contributions of our work can be summarised as follows:
\begin{itemize}
    \item We propose a novel view on the unsupervised discovery of geometrically meaningful landmarks that, instead of relying on proxy or auxiliary losses, uses a self-training strategy that refines an initial set of unindexed keypoints to endow them with geometrically-aware descriptors. 
    \item To the best of our knowledge, our approach, which alternates between correspondence recovery for pseudo-labelling and a contrastive loss for feature learning, is the first to directly propose a geometrically aware objective for unsupervised discovery through pseudo-labelling.
    \item Contrary to previous works, our method can deal with viewpoint changes thanks to an over segmentation of the feature space that accounts for viewpoint-specific descriptors of the same landmark. 
    \item We conduct extensive ablation studies and deliver competitive results in various challenging tasks and object categories.
\end{itemize}

\section{Related Work}
This paper brings the reasoning behind \textit{clustering algorithms} for self-supervised representation learning to iteratively refine \textit{generic keypoints}, and endow them with semantic meaning, in a process commonly known as \textit{unsupervised landmark discovery}. As such, we provide a brief review on these three topics, departing from the latter, as it constitutes the main goal of this paper. \\

\textbf{Landmark Discovery.} Our goal in this paper is to build a landmark detector $\Psi$ that can be learned without human supervision. Landmarks convey semantic information about a particular object and serve the modelling of rigid and non-rigid deformations. Because of this, a landmark detector must be \textit{equivariant} to geometric transformations $g$, i.e. if an image $\textbf{x}$ undergoes an image deformation defined by $g(\textbf{x})$, the detector must follow suit: $\Psi(g(\boldsymbol{x})) = g(\Psi(\boldsymbol{x}))$. Such a simple yet essential requirement was the driving force behind the first method on unsupervised landmark discovery~\cite{Thewlis2017UnsupervisedLO}, where a network is trained to produce $K$ heatmaps from which the corresponding landmark locations are derived through a differentiable \textit{softargmax} operator~\cite{Yi2016LIFTLI}. By imposing the equivariant constraint on images and known deformations, as well as by adding auxiliary losses to avoid trivial solutions, the network can discover a set of $K$ meaningful landmarks. The concept of equivariance can also be extended and used to learn networks that are designed to output dense feature maps rather than heatmaps~\cite{Thewlis2017UnsupervisedLOobjectframes}. While such extension does not aim at ``discovering" object landmarks, it is possible to learn, on a few-shot basis, a per-landmark regressor, i.e. a regressor from feature maps to landmarks from a handful set of annotated samples. A similar approach was also extended to learn object symmetries without regard to the specific task of landmark discovery~\cite{Thewlis2018ModellingAU}. The equivariance constraint was also used to learn dense feature representations that cope with intra-class variation by exchanging features~\cite{Thewlis2019UnsupervisedLO} between images before applying equivariance.

The use of equivariance as a proxy task to learn landmark detectors is usually prone to finding landmarks that do not have a proper semantic meaning (e.g. in the background). To avoid this issue, a different alternative consists of considering the proxy task of \textit{image generation}, whereby a landmark detector is a necessary intermediate step to capture the geometry of an object for a decoder to generate a version of the input image~\cite{jakab2018unsupervised, NIPS2019_9505}. These frameworks share a common structure, consisting of a landmark detector, a ``geometry distillation" bottleneck, and a conditional image generator. The detector and the bottleneck are meant to represent the object's geometry, which is forwarded to the conditional image generator along with a deformed version of the image. The whole pipeline is trained end-to-end with an image reconstruction loss. An alternative version~\cite{Zhang_2018_CVPR} advocates for a differentiable autoencoder framework. Similar methods have also appeared, combining both equivariance and image generation for object feature representation~\cite{Kulkarni2019UnsupervisedLO, Siarohin2019AnimatingAO, Shih2019VideoIA, cheng2020unsupervised}, or attempting to disentangle pose from appearance~\cite{Shu2018DeformingAU,Lorenz_2019_CVPR}, which do not explicitly aim at learning object landmarks. These methods also suffer from the drawback of not being explicitly designed to produce semantically meaningful landmarks. On the contrary, our framework sets a novel direction whereby generic keypoints are transformed into semantically meaningful landmarks. \\

\begin{figure*}
\centering
\includegraphics[width=\linewidth]{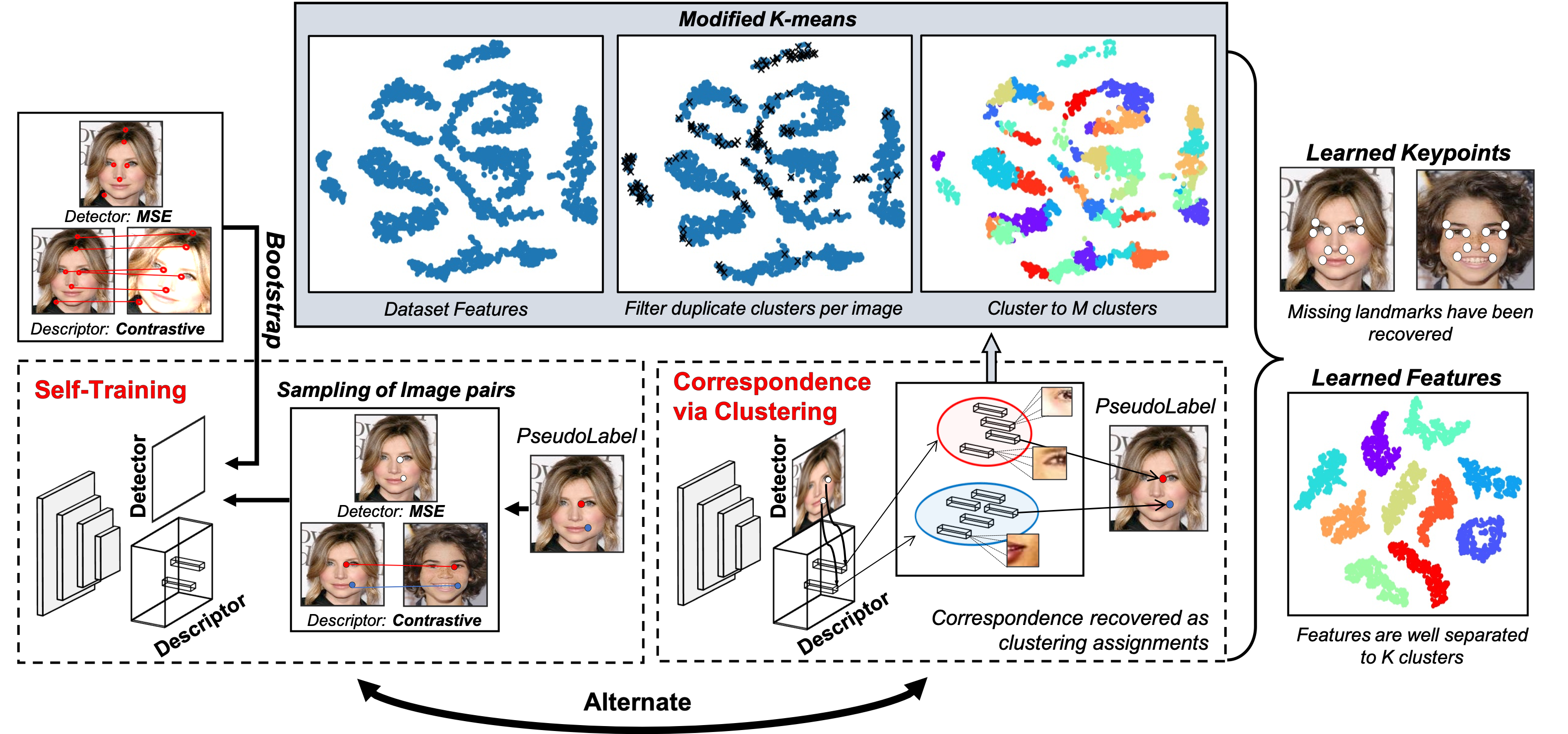}
\caption{Stage 1 of our proposed framework. A neural network is learned with two separate output heads (detector and descriptor head). During training, we alternate between correspondence recovery via clustering and self-training using the recovered correspondences. Training is bootstrapped by generic keypoints. In contrast to recent approaches, our framework enables learning of local features from unpaired image data. Correspondence is recovered via clustering following our Modified-KMeans algorithm. Our method is able to recover missing landmark locations and converge to well-separated features that can be used for accurate correspondence recovery. Dataset feature visualisation created through t-SNE\cite{Maaten2008VisualizingDU}.}
\label{fig:pipeline}
\end{figure*}

\textbf{Keypoint detection.} Keypoints, also known as salient or fiducial points, are used to represent the locations in an image that are of interest without regard to any semantic meaning. Keypoint detection is a critical step for any sparse image matching algorithm (Structure-from-Motion, Simultaneous Localisation and Mapping, 3D reconstruction, etc). Keypoints are accompanied by descriptors that allow their matching across different images, i.e. that allow \textit{correspondence recovery}. Early works in keypoint detection and description were primarily based on computing local image variations, such as the histograms of the magnitude and orientation of image gradients (e.g. HOG~\cite{Lowe1999ObjectRF}, SIFT~\cite{Lowe2004DistinctiveIF}, SURF~\cite{Bay2006SURFSU}, and variants~\cite{Miao2013InterestPD,Alcantarilla2012KAZEF,Salti2013KeypointsFS}) or the binary comparisons between neighbouring pixels(e.g. LBP~\cite{Pietikinen2016TwoDO}).

Lately, there is an increasing interest in ``learning" keypoint detectors and descriptors, using CNN-based approaches that can produce dense features~\cite{Lenc2016LearningCF,DeTone2018SuperPointSI,Revaud2019R2D2RA}. Given that (in most cases) there is no concept of ``ground-truth" keypoints, learning-based approaches work on an unsupervised setting, defining a proper proxy or auxiliary objective, e.g. invariance to viewpoint changes~\cite{Lenc2016LearningCF,Yi2016LIFTLI}, or feature discriminativeness~\cite{Revaud2019R2D2RA}. In this paper, we study the feasibility of the keypoints detected by some of these methods to be converted into landmarks, observing that the strongest initialisation comes from those given by SuperPoint~\cite{DeTone2018SuperPointSI}, which uses a three-stage approach with synthetic pre-training, homographic recovery, and discriminative matching. \\

\textbf{Self-training via clustering.} Self-training refers to a set of methods where a model's own predictions are used as pseudo-labels for model training. Common methods for self-training can include converting the highly confident predictions into hard-labels~\cite{Sohn2020FixMatchSS,Xie2019SelftrainingWN}, the opposite~\cite{Rizve2021InDO}, or applying a model ensemble~\cite{Nguyen2020SELF}. Most self-training approaches focus on the task of image classification~\cite{Sohn2020FixMatchSS,Xie2019SelftrainingWN,Rizve2021InDO} whereby each training image is considered a particular class. Self-training is also applied for unsupervised segmentation~\cite{Dai2015BoxSupEB, Khoreva2017SimpleDI}, foreground-background segmentation~\cite{Faktor2014VideoSB,Stretcu2015MultipleFM} and salience object detection~\cite{Zhang2017SupervisionBF}. 

A recent line of methods for self-training relies on the concept of \textbf{clustering} to generate pseudo ground-truth annotations~\cite{Caron2018DeepCF,Noroozi2018BoostingSL,Li2016UnsupervisedVR,Zhuang2019LocalAF,Yan2020ClusterFitIG,huang2018and,asano2020self,caron2020unsupervised}. These approaches are based on computing a set of clusters that can be used to ``label" the training images. An optimisation objective can be derived from these pseudo-labels, e.g. the typical cross-entropy~\cite{Noroozi2016UnsupervisedLO, Caron2018DeepCF}, a cluster identification~\cite{Li2016UnsupervisedVR}, or even optimal transport problem \cite{asano2020self,caron2020unsupervised}. Related are also methods that utilise clustering-based interfaces for feature grouping. The slot-attention mechanism~\cite{Locatello2020, Kipf2022ConditionalOL} computes a set of exchangeable representations or slots (analogous to cluster centroids), that can bind to any object of a complex input scene. VQ-VAE and VQ-VAE-2~\cite{Oord2017NeuralDR,Razavi2019GeneratingDH} propose an unsupervised approach for learning local representations by mapping image patches into a categorical distribution of latent, learnable embeddings. The encoder assigns a cluster centroid to each image patch, and the decoder is targeted with reconstructing the input image from the grid of selected centroids. Routing mechanisms based on soft-clustering are also used by the SetTransformer in \cite{Lee2019SetTA}, for processing set-structured data through multiple attention heads (for each cluster assignment), as well as for training a generalized VLAD layer in \cite{Arandjelovi2018NetVLADCA}. In all these cases, the ultimate goal is to learn a network that produces strong feature representations in an unsupervised setting to be applied to a downstream task thereafter. The pseudo-labels are not defined to convey a meaning that relates to the downstream task, and are generally discarded after training. To our knowledge, our work is the first to apply self-training with automatically generated pseudo-labels.

\section{Method}
\label{sec:method}

\def\Im{\mathcal{I}}
\def\Loss{\mathcal{L}}
\def\x{{\bf x}}
\def\f{{\bf f}}
\def\c{{\bf c}}
\def\p{{\bf p}}
\def\y{{\bf y}}
\def\s{{\bf s}}
\def\Re{\mathbb{R}}
\def\prob{\mathbb{P}}
\def\hIm{\hat{\mathcal{I}}}
\def\hI{\hat{I}}
\def\eE{\mathbb{E}}
\def\X{\boldsymbol{\mathcal{X}}}
\def\Y{\boldsymbol{\mathcal{Y}}}
\def\F{\boldsymbol{\mathcal{F}}}
\newcommand{\net}[1]{\ensuremath{\Phi(#1)}}
\newcommand{\bb}[1]{\bf{#1}}
\newcommand{\h}[1]{\tilde{#1}}
\newcommand{\bbold}[1]{{[\bf #1]}}
\newcommand{\obold}[1]{{\bf #1}}

\subsection{Problem statement} \label{ssec:problem}

Let $\mathcal{X} = \{{\bf x} \in \Re^{W \times H \times 3}\}$ be a set of $N$ images of a specific object category (e.g. faces, human bodies etc.). After running a generic keypoint detector on $\mathcal{X}$, our training set $\mathcal{X}$ becomes $\{\x_j, \{\p^j_i\}_{i=1}^{N_j}\}$, where $\p^j_i \in \Re^{2}$ is a keypoint and $N_j$ the number of detected keypoints in image $\x_j$. The original keypoints $\p^j$ for the $j$-th image are not ordered or in any correspondence with object landmarks. Also, multiple object landmarks will not be included in $\p^j$. Finally, some keypoints will be outliers corresponding to irrelevant background. Using only $\mathcal{X}$, our goal is to train a neural network $\mathbf{\Psi}: \mathcal{X}\rightarrow \mathcal{Y}$, where $\mathcal{Y} \in \Re^{H_o \times W_o \times K}$ is the space of output heatmaps representing confidence maps for each of the $K$ object landmarks we wish to detect. Note that the structure of $\mathcal{Y}$ implies that both order and landmark correspondence is recovered.

We will break down our problem into two stages. In the first stage, we will train a network $\mathbf{\Phi}$ producing a set of keypoints with landmark-aware descriptors, which aims to establish landmark correspondence, recover missing object landmarks and filter out irrelevant background keypoints. Then, we will use the output of this stage to train $\mathbf{\Psi}$ in a ``supervised" way, using the pseudo-labels produced by $\mathbf{\Phi}$. Sections \ref{ssec:keypoints2landmarks},~\ref{subsec:corerec},~\ref{ssec:losses} and ~\ref{ssec:warmup} are devoted to describing the first stage (\textbf{Stage1}) of our method, also depicted in Fig.~\ref{fig:pipeline}. Section \ref{ssec:learning_detector} describes the second stage (\textbf{Stage2}), and Section \ref{ssec:flipping} introduces our flipping augmentation strategy. 

\subsection{Network Architecture} \label{ssec:keypoints2landmarks}

Our first stage comprises learning a network $\mathbf{\Phi}$ in a similar fashion to those of keypoint detectors, with a shared backbone $\mathbf{\Phi}_{b} : \mathcal{X} \rightarrow \mathcal{F}$ producing a set of intermediate features $\mathcal{F}$ and two heads: one for detecting the object landmarks $\mathbf{\Phi}_{d}$ and one for landmark-distinctive feature descriptor $\mathbf{\Phi}_{f}$. 

The \textbf{detector head} $\mathbf{\Phi}_d$ will produce, for image $\x_j$, a single-channel spatial confidence map $H_j = \mathbf{\Phi}_d( \mathbf{\Phi}_b( \x_j)) \in \Re^{H_o \times W_o \times 1}$ representing the presence/absence of an object landmark at a given location, without regard to any order or correspondence. We use non-maximum suppression to extract from $H_j$ the landmark locations $\p^j_i$. The main purpose of $\mathbf{\Phi}_{d}$ is to recover the originally missed object landmarks, as well as to assign to each subsequent pseudo-label a corresponding spatial location.

The \textbf{feature extractor head} $\mathbf{\Phi}_f$ will produce for image $\x_j$ a dense feature map $\mathbf{F}_j = \mathbf{\Phi}_f (\mathbf{\Phi}_b( \x_j)) \in \Re^{H_o \times W_o \times d}$ that will be used for \textbf{recovering correspondence}. At each landmark position $\p^j_i$ activated by the detector head, we will extract a $d$-dimensional feature descriptor $\f^j_i$ from $\mathbf{F}$. We use local features for recovering the correspondence of each individual keypoint through clustering.

\subsection{Correspondence recovery}
\label{subsec:corerec}
After applying $\mathbf{\Phi}$ on the training set, $\mathcal{X}$ becomes  $\{\x_j, \{\p^j_i,\f^j_i\}_{i=1}^{N_j}\}$. Then, our first step in the iterative algorithm becomes using the features $\f$ to assign each keypoint a pseudo-label. We refer to this operation as correspondence recovery, as it allows us to identify correspondence of object parts across different images. To assign each detected keypoint a pseudo-label, we follow \cite{Caron2018DeepCF} and perform K-means clustering on the collection of features $\f$. However, different from \cite{Caron2018DeepCF} where the clusters are used to make similar images have similar descriptors in an unsupervised way, our cluster assignment is indeed assigning a meaning label to a given keypoint. For this reason, we observe that it is important not to assign two different keypoints on a given image to the same cluster.

The clustering operation is then defined as:

\begin{multline}
\min_{{\bf C} \in \mathbb{R}^{d \times M}} \frac{1}{N} \sum_{i=1}^N \sum_{j=1}^{N_j} \min_{{\bf y}^j_{i} \in \{0,1\}^M} \| \f_i^j - {\bf C} {\bf y}^j_{i} \|^2_2 \;\;\; \\
\text{s.t.} \;\;\; {\bf 1}_M^T {\bf y}^ j_{i}  = 1 \hspace{5pt} \text{and} \hspace{5pt}
\|\sum_{j} {\bf y}^j_{i}\|_0 = N_j, 
\label{eq:magic_clustering}
\end{multline}

where $M$ is the number of clusters, $\y^j_i$ is the cluster assignment for landmark $\p^j_i$, $\boldsymbol{C}$ is the $d \times M$ centroid matrix and ${\bf 1}_M = [1, \dots ,1]$ is an $M$-d column vector with all entries set to $1$. While in \cite{unsupervLandm2020} the cluster assignment was performed using the Hungarian algorithm~\cite{Kuhn1955TheHM}, here we opt for a simpler solution that attains similar results. For a given image $j$, we find $\{{\bf y}^j_{i}\}$, $\{\f^j_i\}$ by simply keeping, for each cluster $k$, the keypoint whose descriptor is closest to the centroid, i.e. we remove duplicate occurrences of the same cluster $k$ on a single image. Enforcing a single keypoint per cluster for each image also provides a natural way of filtering out noisy keypoints. Given that a keypoint with a more representative feature has already been found for a cluster $k$ in a particular image, it is likely that the second occurrence would be a noisy point. 

While in \cite{unsupervLandm2020} the number of object landmarks was automatically discovered after progressive merging of similar clusters, here we enforce the detection of at most $K$ clusters per image, in accordance with other recent unsupervised landmark detectors~\cite{Thewlis2017UnsupervisedLO,Zhang_2018_CVPR,jakab2018unsupervised,NIPS2019_9505}. We do that by additionally constraining $\mathbf{\Phi}_{d}$ to detect at most $K$ keypoints per image (one per detected cluster). To that end, the modified K-means algorithm is executed \textit{twice}: (1) the first time, clustering is performed with $M=K$, to filter out duplicate occurrences of the same cluster in a single image and constrain our training set to at most $K$ points per image (the detection of less than $K$ keypoints is allowed due to factors like occlusion). Note that this clustering step is solely performed as a filtering mechanism and produced pseudolabels ${\bf y}^j_{i}$ are discarded.

To calculate the final ${\bf y}^j_{i}$'s we (2) cluster the reduced set of features a second time with $M \gg K$. Setting a larger $M$ forces the clustering algorithm to split the $K$ underlying landmark classes into multiple smaller clusters leading to the formation of multiple clusters capturing the same underlying landmark. This is similar to \cite{Caron2018DeepCF} where the best performance is obtained by clustering the $1,000$ ImageNet classes to $10,000$ clusters during pseudolabel formation. The resulting over-segmentation of the feature space is necessary for cases where viewpoint changes introduce significant appearance changes. This differentiates our approach from prior works, which do not account for large out-of-plane rotations. An illustration of this in the form of a t-SNE\cite{Maaten2008VisualizingDU} visualisation is shown in Fig. \ref{fig:pipeline}. Even though clustering is performed twice, this step can be executed fast by using an accelerated similarity search method~\cite{JDH17}.

\begin{figure}
\centering
\includegraphics[width=0.9\linewidth]{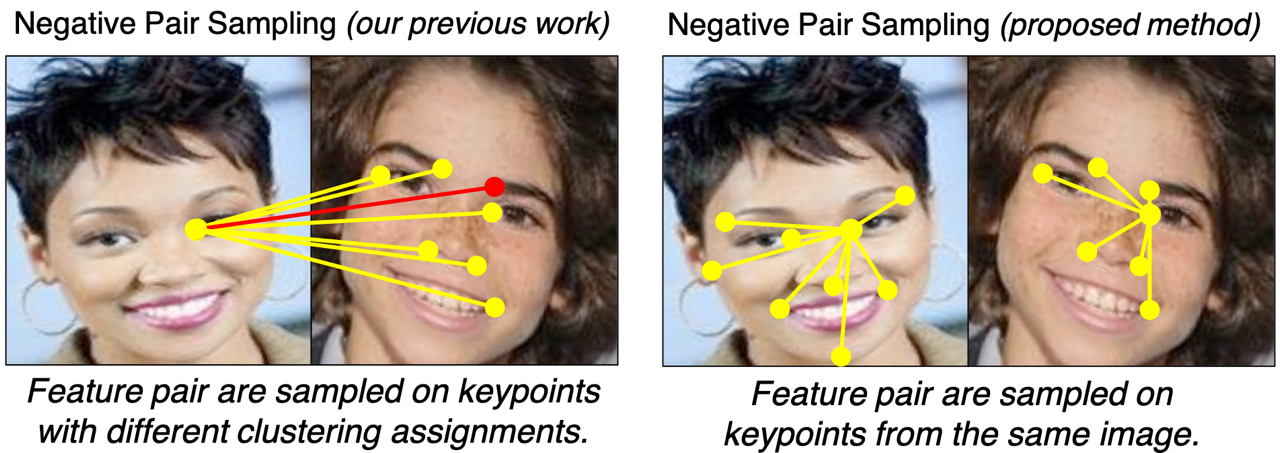}
\caption{Proposed negative pair mining strategy compared to \cite{unsupervLandm2020}. In \cite{unsupervLandm2020}, negative pairs are sampled on keypoint locations with different clustering assignments. Since multiple clusters can track the same landmark, this can lead to inaccurate negative pairs \textit{(red line)}. Sampling negatives from the same image ensures accurate pairs given that by definition, each landmark can only appear once per image. }
\label{fig:pairs}
\end{figure}

\subsection{Training Losses}
\label{ssec:losses}
After the correspondence recovery step described in Sec.\ref{subsec:corerec}, the training set has now been augmented to include two different sets of pseudo-labels: the keypoint positions $\p_j$ and the corresponding cluster assignments $\y_j$. The next step consists then of training the network $\Phi$, with both its backbone $\Phi_b$ and heads $\Phi_d$ and $\Phi_f$, using the generated pseudo-labels. At the end of this step, the training set will be re-populated with the network's output: a new set of keypoints and descriptors will be generated, and new clustering assignments will be calculated.

The loss corresponding to the \textbf{detector head} is the standard MSE loss, defined as 
\begin{equation}
    \mathcal{L}_d(\x_j) =  \| H(\mathbf{x}_j) - \mathbf{\Phi}_{d}(\mathbf{\Phi}_b(\mathbf{x}_j)) \|^2,
\label{eq:mse}
\end{equation}
where the ground-truth heatmap $H$ for a given image $\mathbf{x}_j$ is formed by placing 2D-Gaussian maps on each of the keypoint locations $\{\p_j^i\}_{i=1...N_j}$. Our self-training approach confirms recent findings~\cite{Arpit2017ACL,Rolnick2018DeepLI} that show that over-parameterized neural networks tend to learn noiseless classes first, before overfitting to noisy labels in order to further reduce the training error. We observe such a pattern in learning object landmarks: a true landmark that commonly appears in the training set results in high detection confidence. Similarly, background locations that do not recurrently follow a specific pattern tend to be filtered out. 

For the \textbf{feature extractor head} we propose the use of a contrastive loss. Note that this differs from \cite{Caron2018DeepCF} which uses a classifier to generate the pseudo-labels and a cross-entropy loss to update the network. Given the augmented training set at some training iteration $t$, $\mathcal{X}_t = \{\x_j, \{\p^j_i,\y^j_i\}_{i=1}^{N_j}\}$ our goal is to update $\Phi_f$ to produce features that, when extracted at some keypoints $\p^i_j$ and $\p^{i'}_{j'}$ for some locations $i,i'$ on images $j$ and $j'$, respectively, are similar if and only if the corresponding pseudo clusters match, i.e. if $\y^j_i = \y^{j'}_{i'}$. To do so, we resort to a contrastive loss, where the goal is to bring pairs of features corresponding to the same cluster close whilst pulling features from different clusters apart. For a given pair of images $\x_j$ and $\x_{j'}$ and output locations $i$ and $i'$, the contrastive loss is formulated as:

\begin{multline}
    \mathcal{L}_{f}(\mathbf{x}^j_i, \mathbf{x}^{j'}_{i'}) = \\ 
    {\bf 1}_{[y^j_i = y^{j'}_{i'}]} \|\f^{j}_i - \f^{j'}_{i'} \|^2 + 
    {\bf 1}_{[y^j_i \neq y^{j}_{i'}]} \text{\footnotesize max}(0,m-
    \|\f^{j}_i - \f^{j}_{i'} \|^2)
\end{multline}
where recall $\f^j_i = \Phi_f(\Phi_b (\mathbf{x}_j))_i$ is the $d$-dimensional feature vector extracted, for image $j$ at the position $\mathbf{p}_i$, from the output of the feature head. A margin $m$ is used to enforce features corresponding to negative pairs to be far apart.

As is common in unsupervised learning methods that build on contrastive learning, the choice of positive and negative pairs plays an important role in the learning process. Positive pairs can now be formed from different images where two keypoints are assigned the same cluster, as well as from two images where one is a synthetic deformation of the other. On the other side, negative pairs can be chosen in many different ways. While in \cite{unsupervLandm2020} the negative feature pairs were selected randomly from the keypoint locations at different images (excluding those for which the pseudo-label was the same), in this work, we improve our negative mining by choosing all the negatives from the same image only. Given the over-segmentation of the underlying landmarks to $M$ clusters, the same landmark in two different images could be assigned to different clusters, which would hinder the learning process. On the other hand, as noted above, each object landmark can only appear once per image. Thus, features extracted at any other location $i'$ far from $i$, even when not corresponding to any proper keypoint $\p$, is a good, informative negative pair. An illustration of our negative pair mining strategy compared to that in \cite{unsupervLandm2020} is show in Fig. \ref{fig:pairs}. 

Denoting by $\theta_b$, $\theta_d$ and $\theta_f$ the parameters of $\mathbf{\Phi}_b$, $\mathbf{\Phi}_{d}$ and $\mathbf{\Phi}_{f}$, respectively, the full training procedure for Stage 1 is summarised in Algorithm~\ref{alg:stage_1}.

\begin{algorithm}[h!]
\DontPrintSemicolon
\SetAlgoLined
\KwData{$\mathcal{X}_0 = \{\x_j, \{\p^j_i\}_{i=1}^{N_j}\}$}
Compute $\y^j_i$ using Eqn.~\ref{eq:magic_clustering}  \;
Set $\mathcal{X}_0 = \{\x_j, \{\p^j_i,\y^j_i\}_{i=1}^{N_j}\}$ \;
\For{$t = 1 : T$}{
\For{$n=1 : N_{iters}$}{
Sample batch \;
$(\theta_b, \theta_d) \leftarrow (\theta_b, \theta_d) -\nabla_{\theta_b,\theta_d} \mathcal{L}_d $\;
$(\theta_b, \theta_f) \leftarrow (\theta_b, \theta_f) -\nabla_{\theta_b,\theta_f} \mathcal{L}_f $\;
}
Update $F$ and $p^j$ using frozen $\Phi$ \;
Compute $\y^j_i$ using Eqn.~\ref{eq:magic_clustering}  \;
Update $\mathcal{X}_t = \{\x_j, \{\p^j_i,\y^j_i\}_{i=1}^{N_j}\}$  \;
 \;

}
\caption{Stage 1 training}
\label{alg:stage_1}
\end{algorithm}

\subsection{Bootstrapping}
\label{ssec:warmup}
Initially, at round $t=0$ the training set  $\mathcal{X}_0$ only includes $\{\x_j, \{\p^j_i\}_{i=1}^{N_j}\}$ without point correspondences $\f^j_i$, needed for correspondence recovery as described in Sec. \ref{subsec:corerec}. In \cite{unsupervLandm2020} the initial features were given by the generic keypoint descriptor, from where an initial clustering step could be performed. In this paper, we opt for a \textit{warm up} pre-training stage where we train the feature extractor using only pairs of images in which one is a synthetic deformation of the other. We form known point correspondences through synthetic augmentations that can be used as initial positive pairs. This corresponds to initialising our backbone and feature extractor head using equivariance.

\subsection{Learning an object landmark detector}
\label{ssec:learning_detector}
At the end of Stage 1, the training set $\mathcal{X}$ is composed of a series of keypoints with landmark-aware descriptors. However, our goal is to train a network that can detect \textit{a fixed number of $K$ landmarks}.

Provided that the training set is now composed of $M \gg K$ clusters, training a landmark detector on $K$ classes is not trivial because it is unknown which clusters correspond to the same landmark. In \cite{unsupervLandm2020} this process was tackled through a progressive merging step that was eventually reducing the number of clusters. However, thanks to the fact that the number of keypoints per image is now limited to $K$, as well as to the negative mining strategy, we observe that the learned features automatically form $K$ well-separated clusters (as can be seen in Fig. \ref{fig:tsne} for $K=30$). This observation thus eliminates the need for a progressive cluster merging step. 

To finally populate our training set with $K$ clusters only, we perform a last K-means clustering with $K$ clusters only. Then, we can train $\Psi$ using standard Heatmap Regression, by placing a Gaussian at the pseudo ground-truth landmarks provided by $\Phi$. Because not every image will be pseudo annotated with $K$ landmarks, we only compute the error for those that are available. For a given image $\mathbf{x}_j$ we assume that the network $\Phi$ has produced a set of $K' \leq K$ indexed landmarks. Then, the index set for the pseudo ground-truth landmarks for image $\mathbf{x}_j$ can be split into the \textit{detected set} $\mathcal{D}_j \subseteq [0,\cdots,K-1]$, and \textit{missed set} $\mathcal{M}_j \subset [0,\cdots,K-1]$ with $\mathcal{D}_j \cap \mathcal{M}_j = \emptyset$. Denoting with $H(\mathbf{x}_j)_{k}$ the $k$-th pseudo ground-truth heatmap for landmark $k$ and with $\Psi(\mathbf{x}_j)_k$ the $k$-th heatmap produced by $\Psi$, the error for image $\mathbf{x}_j$ is defined as:
\begin{equation}
E(\mathbf{x}_j | \Psi) = \frac{1}{|\mathcal{D}_j|} \sum_{k \in \mathcal{D}_j} \| H(\mathbf{x}_j)_k - \Psi(\mathbf{x}_j)_k \|^2
\end{equation}
where $|\mathcal{D}_j|$ denotes the number of detected landmarks. Because the discovered landmarks appear with high frequency on the training set, the network $\Psi$ not only learns to correct the possible outliers of $\Phi$, but will also ensure a set of $K$ landmarks is detected. 

\subsection{Flipping augmentation}
\label{ssec:flipping}
Flipping is a common augmentation strategy when training a landmark detector. In the supervised case, one can flip an image and mirror the ground-truth landmarks, given the naturally known correspondence between landmarks and their mirrored counterparts. In the unsupervised learning case, such correspondence is not known. In methods based on generative modelling or equivariance, one can only resort to flipping both the original and the synthetically generated image. This paper proposes to recover the symmetric landmark correspondences using clustering. At the correspondence recovery step (Sec. \ref{subsec:corerec}), pairs of features are sampled on both an image and its flipped version. We treat these features independently and produce 2 cluster assignments for each keypoint (one for the original and one for the flipped image). During the training of Stage 1, the cluster assignments of the flipped features are used when an image is randomly flipped. For Stage 2, we find cluster symmetries by measuring maximal correspondence between clusters in the original and flipped images over the whole dataset. Note that in Stage 2, flipping can be used both in training and test time as usually done with supervised landmark detectors.

\section{Experimental Details}

We first begin with describing the employed datasets (Sec.~\ref{ssec:datasets}) as well as the general implementation details (Sec.~\ref{ssec:implementation}). We then analyse the different parts of our method in Sec.~\ref{sec:ablation}, and compare the performance of our approach w.r.t. competing methods in Sec.~\ref{sec:sota}. 

\subsection{Datasets}
\label{ssec:datasets}

\textbf{Facial datasets}. We evaluate our method on the commonly used \textbf{CelebA-MAFL}~\cite{Liu2015DeepLF, Zhang2014FacialLD} and \textbf{AFLW}~\cite{Kstinger2011AnnotatedFL} datasets, as well as on the challenging \textbf{LS3D}~\cite{Bulat2017HowFA}. The \textbf{CelebA} dataset contains $\sim 200$K facial images manually annotated with $5$ facial landmarks. We follow prior work and remove from the training the $1000$ images corresponding to the MAFL partition~\cite{Zhang2014FacialLD} which is used for evaluation. We extract loose crops around the target objects using facial bounding boxes. Since these are not provided we precompute them using \cite{Zhang2017S3FDSS}. A small margin is added in each direction (of the bounding box) similar to \cite{NIPS2019_9505}. Further details about the preprocessing can be found in \footnote{\url{https://github.com/dimitrismallis/KeypointsToLandmarks}}. The \textbf{AFLW} contains $10,112$ training images and $2,991$ test images annotated with 21 landmarks. We use the same partition as \cite{jakab2018unsupervised,NIPS2019_9505} and directly rescale the images to $256 \times 256$ without performing additional cropping (they are already tightly cropped). Both CelebA and AFLW are annotated with a limited number of points which in practice limit the evaluation of unsupervised methods to capture proper geometric deformations. For this reason, we opt for re-annotating both datasets with $68$ landmarks using the 2D detector of \cite{Bulat2017HowFA}. We evaluate both our and competing methods using the same set of detected points. The \textbf{LS3D}~\cite{Bulat2017HowFA} dataset contains images of faces with large pose variations. It is constructed by re-annotating the images from 300W-LP~\cite{zhu2016face},  AFLW~\cite{Kstinger2011AnnotatedFL}, 300VW~\cite{Shen2015TheFF}, 300W~\cite{Sagonas2013300FI} and FDDB~\cite{Jain2010FDDBAB} in a consistent manner with $68$ points using the automatic method of \cite{Bulat2017HowFA}. We extract loose crops around the
target objects using the provided facial bounding boxes. A small margin is added on each direction similar to CelebA preprocessing. Note that LS3D dataset is annotated with 3D points. Evaluation is performed on the LS3D-W Balanced test set, comprising $7200$ images, including an equal number of images for each of the range of yaw angles $[0^o-30^o]$, $[30^o-60^o]$, $[60^o-90^o]$.

\textbf{Human Body datasets}. We evaluate our method on \textbf{BBCPose}~\cite{Charles2013DomainAF}, \textbf{Human3.6M}~\cite{Ionescu2014Human36MLS} and \textbf{PennAction}~\cite{Zhang2013FromAT}. \textbf{BBCPose}~\cite{Charles2013DomainAF} is a dataset of $20$ sign language videos ($10$ for training, $5$ for validation and $5$ for testing) annotated with $7$ human pose landmarks (head, wrists, elbows, and shoulders). We form the training set by selecting 1 of every 10 frames leading to a set of $60885$ images. Evaluation is performed on the standard test set ($1000$ images). \textbf{Human3.6M}~\cite{Ionescu2014Human36MLS} is an activity dataset with a constant background containing videos of actors in multiple poses under different viewpoints. We follow the evaluation protocol of \cite{Zhang2018UnsupervisedDO} and use all $7$ subjects of the training set ($6$ subjects were used for training and $1$ for testing) on six activities (direction, discussion, posing, waiting, greeting, walking). We form our training set by extracting 1 every 50 ($48240$ training images) and 1 every 100 frames for testing ($2760$ images). Contrary to \cite{Zhang2018UnsupervisedDO} we do not perform background subtraction to simplify landmark detection. \textbf{PennAction}~\cite{Zhang2013FromAT} is a dataset of 2326 videos of humans participating in sports activities. For this experiments, we use the same 6 categories as in \cite{Lorenz2019UnsupervisedPD} (tennis serve, tennis forehand, baseball pitch, baseball swing, jumping jacks, golf swing). For this experiment, we do not use the provided $50\% - 50\%$ train-test split to ensure sufficient training data. We opt for using the 5 first videos for each category to form a separate test set. This results in $51661$ training and $1776$ testing images. For all human pose datasets, we crop images using each person's bounding box. Since BBCPose bounding boxes are not provided, we compute them using the popular detector of \cite{Ren2015FasterRT}. The scale of each subject is set to $s=c*H/200$, where $H$ is the height of the bounding box. For BBCPose and Human3.6, $c$ is set to $1.1$ (similar to \cite{xiao2018simple}). For PennAction we set $c=1.35$.

\textbf{Other datasets}: In addition to the above categories, we also evaluate our method on the \textbf{Cat Heads}~\cite{Zhang2008CatHD} dataset, which consists of 9k images of cat heads annotated with 9 landmarks. We use the test-train split of \cite{Zhang2018UnsupervisedDO} with $7747$ training and $1257$ testing images. The bounding box is calculated as the tightest box around the ground truth keypoints plus a small margin (similar to CelebA preprocessing). Finally, we present a qualitative evaluation in the \textbf{CUB-200-2011}~\cite{WelinderEtal2010} dataset, which contains $11778$ images of birds belonging to $200$ species. We use the same setting as \cite{Lorenz2019UnsupervisedPD} and remove the seabird species. We extract crops using the provided bounding boxes plus a small margin on each side as previously.

\subsection{Implementation Details}
\label{ssec:implementation}

\textbf{Network architecture:} We use the Hourglass architecture of~\cite{Newell2016StackedHN} with the residual block of~\cite{Bulat2017BinarizedCL} for both $\mathbf{\Psi}$ and $\mathbf{\Phi}$. The image resolution is set to $256 \times 256$. For network $\mathbf{\Phi}$, the localisation head produces a single heatmap with resolution $64 \times 64$, and the descriptor head produces a volume of $64 \times 64 \times 256$, i.e. a volume with the same spatial resolution containing the $256$-d descriptors. The network $\mathbf{\Psi}$ produces a set of $K$ heatmaps, each $64\times64$.

\textbf{Training:} Keypoints are initially populated by SuperPoint~\cite{DeTone2018SuperPointSI}. Before the training starts, we apply an automatic outlier removal step to filter out keypoints most likely to be of no use. We use the Faiss library~\cite{JDH17} for this preliminary step, as well as for the K-means clustering. We perform warm-up for $30,000$ iterations as described in \ref{ssec:warmup}. Then, we apply clustering and update the pseudo-ground truth every $5,000$ iterations. The number of clusters $M$ is set to $100$ for all datasets. A margin $m$ of $0.8$ is used for the constrastive loss $\mathcal{L}_f$ and the two losses are balanced by adding a $\lambda=0.1$ factor to $\mathcal{L}_d$. The algorithm takes around 200,000 iterations to converge in all datasets. For Stage 2, we initialise the model $\Psi$ from the weights of the model $\Phi$ resulting after Stage 1, except for the weights of the last layer that are trained from scratch. To train the models, we used RMSprop~\cite{Graves2013GeneratingSW}, with learning rate equal to $2 \cdot 10^{-4}$, weight decay $10^{-5}$ and batch-size 16. All models were implemented in PyTorch~\cite{paszke2017automatic}. Similarly to other recent methods~\cite{jakab2018unsupervised,Zhang2018UnsupervisedDO}, we also boost the training on video datasets by adding temporal supervision. To that end, the image pairs used for contrastive learning are formed either by randomly sampling two frames from different training videos, or by selecting two nearby frames (with a probability of $0.5$). Given an image $\x_j$ with keypoints $\p^j$, we compute the corresponding keypoints $\p^{j'}$ of an adjacent frame $\x_{j'}$ through sparse optical flow calculation where $\p^{j'}_{i}$ corresponds to $\p^{j}_{i}$ for every $i \in [1,N_j]$  (similar to \cite{Zhang2018UnsupervisedDO}). Positive pairs are formed as the descriptors extracted from corresponding keypoint locations $(\p^{j'}_{i},\p^{j}_{i})$. Notably, our approach achieves good performance without temporal supervision (optical flow is used only when explicitly stated).

\textbf{Evaluation:} Quantitative evaluation of unsupervised landmark detectors is often assessed by quantifying the degree of correlation between manually annotated landmarks and those detected by the proposed approach. This is accomplished by learning a simple regressor with no bias that maps the discovered landmarks to those manually annotated, using a variable number of images in the training set. Numerical evaluation is often measured by means of the Normalised Mean-squared Error (NME). In addition, we follow~\cite{NIPS2019_9505}, and complement this measure (herein referred to as \textbf{Forward-NME}) by measuring the error on a reverted regressor, i.e. one that maps the manual annotations into the discovered landmarks. As found by \cite{NIPS2019_9505}, this measure, known as \textbf{Backward-NME}, helps identify unstable landmarks. We also present Cumulative Error Distribution (CED) curves for these metrics, which permit a per-landmark comparison w.r.t. state-of-the-art methods. We use interocular distance to normalise errors in facial datasets (CelebA, AFLW and CatHeads), and shoulder distance for human pose datasets (BBCPose and Human3.6). Due to the large pose variation on LS3D and PennAction datasets, we opt for normalising the errors using the squared root of the bounding box area, where the bounding box is defined as the smallest rectangle that fits the ground-truth points. 

We are also interested in assessing the quality of the discovered landmarks after Stage 1. Because not all landmarks will be activated in each image after Stage 1, we need to complete the missing values before being able to compute the aforementioned metrics. To do so, we gather all discovered landmarks in a matrix $X \in \mathbb{R}^{K \times N}$, with $N$ the number of training images and $K$ the number of discovered landmarks, and use the Singular Value Thresholding method for Matrix Completion~\cite{cai2010singular}, leaving the detected points unchanged. At test time, we fill the missing landmarks with their corresponding mean positions, computed from the training set. Note that Matrix Completion is used solely for evaluation purposes and not as part of our proposed training framework.

\begin{figure}
\centering
\includegraphics[width=0.9\linewidth]{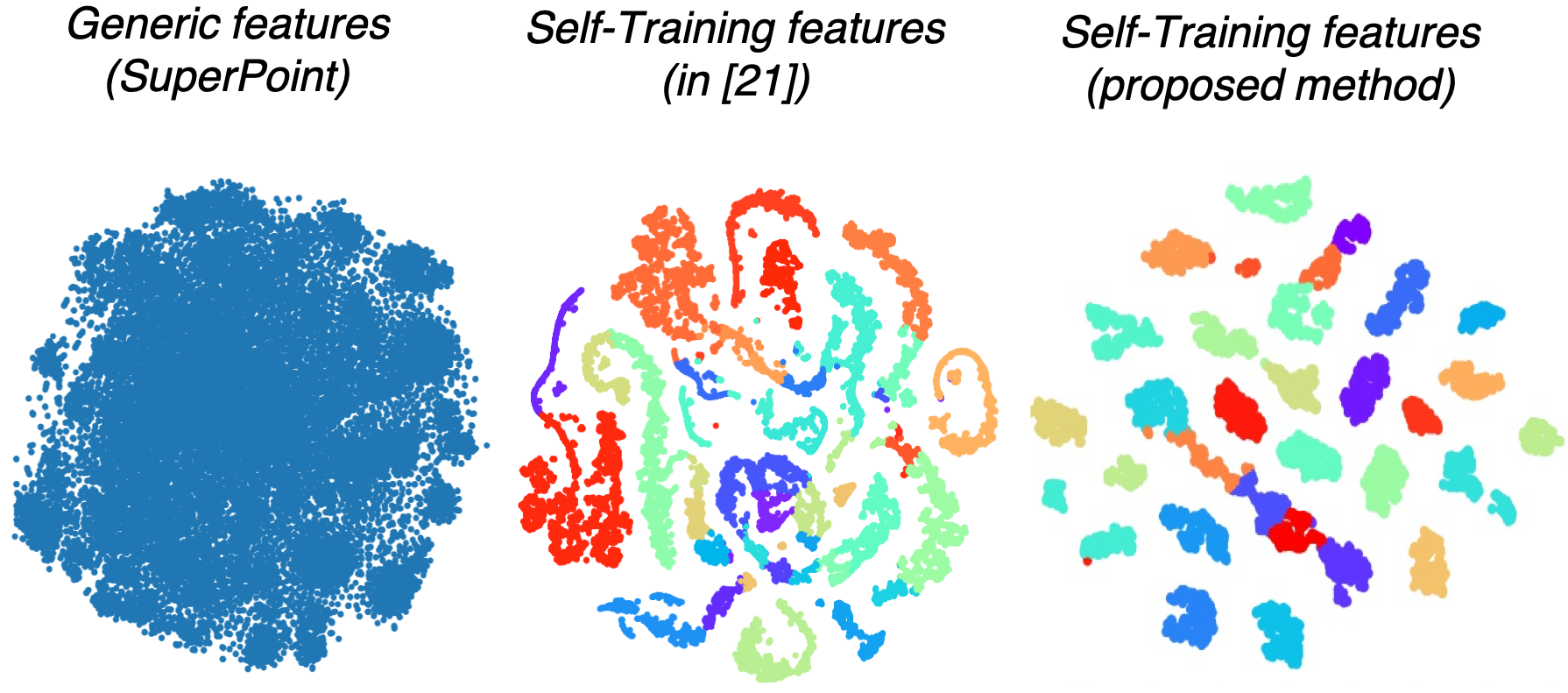}
\caption{t-SNE\cite{Maaten2008VisualizingDU} visualisation of local features. Our new algorithm produces more defined clusters compared to those produced by SuperPoint\cite{DeTone2018SuperPointSI} and our previous work \cite{unsupervLandm2020}.}
\label{fig:tsne}
\end{figure}

\section{Ablation Studies}
\label{sec:ablation}
We perform a series of ablation studies to evaluate different aspects of our proposed method. In particular, we are interested in measuring how the initial conditions affect the training of our proposed approach, as well as the impact of the training components introduced in our method.

\subsection{On the initial conditions}

\begin{figure}
\centering
\includegraphics[width=0.65\linewidth]{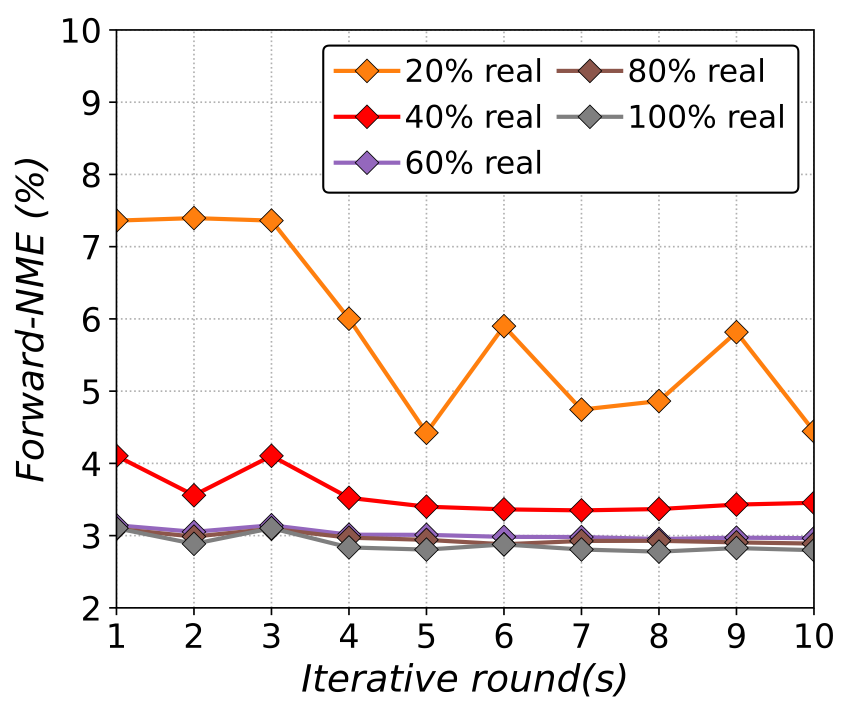}
\caption{Forward-NME (shown for the first $10$ iterative rounds) of training the first stage of our method with varying ratios of real and random points. Experiment is performed on CelebA \cite{Liu2015DeepLF}. Real points are sampled from 15 facial landmarks and further perturbed spatially by a small offset sampled from $[-3px,+3px]$. }
\label{fig:noise}
\end{figure}

\textbf{Robustness to noise} We are firstly interested in measuring to which extent our method can recover semantic correspondence from noisy initialisations. A good initialisation is expected to have some consistent keypoints that overlap to some extent with proper landmarks; a huge number of random keypoints will hinder the learning of landmark correspondence. To evaluate such impact, we first conduct an experiment with synthetic initialisations, i.e. by initialising our training set with a mixture of ground-truth landmark locations and noisy points randomly sampled from the image domain. In particular, we populate each image with a set of 15 points that are either sampled from the ground-truth locations of 15 facial landmarks (eyes, eyebrows, nose, mouth, chin) or chosen at random, uniformly distributed over the image space. Our model is trained to detect 15 object landmarks, and we conduct experiments with varying mixture ratios to evaluate the effect of different noise levels. Fig. \ref{fig:noise} shows the result of this experiment in terms of forward error. Interestingly we find that even with as much as only $20\%$ of real object landmarks in the keypoint initialisation, our method can still perform reasonably well. Increases in the percentage of real points over $40\%$ only result in a slight error reduction. 

\textbf{Keypoint initialisation:} We now evaluate the dependency of our method on initialisations as provided by real keypoint detectors. To this end, we compare the performance of our method, both by means of forward and backward errors, for the case where the initial keypoints are provided by SuperPoint~\cite{DeTone2018SuperPointSI}, R2D2 \cite{Revaud2019R2D2RA}, SIFT \cite{Lowe2004DistinctiveIF} and ORB \cite{Rublee2011ORBAE}. Note that all these methods either are trained in an unsupervised manner (SuperPoint, R2D2), or do not even require training (SIFT, ORB), i.e. neither the initialisation nor our method require any manual supervision. Given that SIFT and ORB tend to detect large numbers of spatially clustered points (that is suboptimal for our purpose of detecting object landmarks), we combine them with Adaptive Non-Maximal Suppression (ANMS~\cite{bailo2018efficient}) to ensure a homogeneous spatial distribution. The results shown in Table~\ref{table:detectorAbl} show that all detectors allow our method to deliver competitive results, with SuperPoint proving to be the overall best choice.

\begin{table}[!t]
\caption{Evaluation in terms of Forward and Backward NME, of landmarks, learned on the first stage of our framework, under different keypoint initialisation methods. Models are trained on CelebA and BBCPose for $K=30$.}\label{table:detectorAbl}
\centering
\begin{tabular}{lcc|cc}
 \toprule
 & \multicolumn{2}{c}{\textit{CelebA}} & \multicolumn{2}{c}
 {\textit{BBCPose}}\\
  \cmidrule{2-5}
 Keypoint Detector  & Fwd & Bwd &  Fwd & Bwd\\
\midrule
    \textit{SIFT} \cite{Lowe2004DistinctiveIF} + \textit{ANMS} \cite{bailo2018efficient} & 4.07 & 7.79 & 21.37 & 15.61\\
    \textit{ORB} \cite{Rublee2011ORBAE} + \textit{ANMS} \cite{bailo2018efficient} & 3.85 & 7.70 & 17.31 & \textbf{11.72}\\
    \textit{R2D2} \cite{Revaud2019R2D2RA} & 3.71 & 7.97 & 18.54 & 12.19\\
    \textit{SuperPoint} \cite{DeTone2018SuperPointSI} & \textbf{3.25} & \textbf{6.65} & \textbf{13.49} & 13.55\\
  \bottomrule
  \end{tabular}
\end{table}

\begin{figure}[!t]
\begin{subfigure}{0.73\linewidth}
\centering
\includegraphics[width=\linewidth]{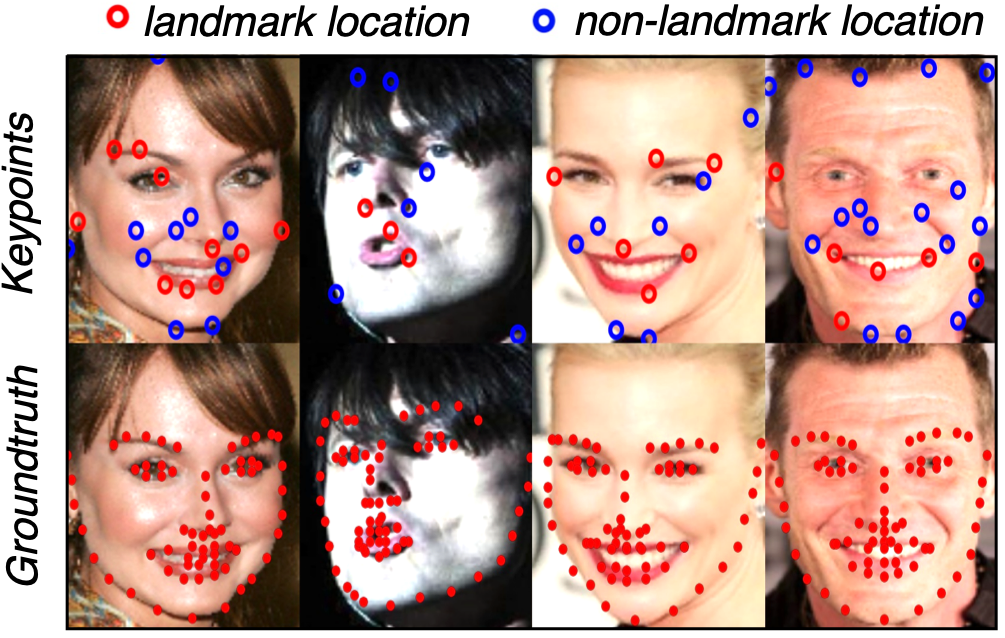}
\end{subfigure}
\vskip 0.25in
\begin{subtable}{\linewidth}
\centering
\begin{tabular}{lc}
\multicolumn{2}{c}{\textbf{Precision ($\%$) w.r.t 68 facial landamrks}}\\
\toprule
 Keypoint Detector  & Precision\\
\toprule
    \textit{SIFT} \cite{Lowe2004DistinctiveIF} + \textit{ANMS} \cite{bailo2018efficient} & 38.9\\
    \textit{ORB} \cite{Rublee2011ORBAE} + \textit{ANMS} \cite{bailo2018efficient} & 47.4 \\
    \textit{R2D2} \cite{Revaud2019R2D2RA} & 54.5 \\
    \textit{SuperPoint} \cite{DeTone2018SuperPointSI} & \textbf{56.5} \\
  \bottomrule
  \end{tabular}
\end{subtable}
\caption{\textbf{(figure-top)} Examples of generic keypoints captured by SuperPoint on facial images along with the corresponding 68 ground-truth landmarks. Generic keypoints capture several object landmark locations (\textit{red keypoints}) as well as non-corresponding background points (\textit{blue keypoints}). \textbf{(table-bottom)} Precision of various generic keypoint detectors w.r.t 68-ground-truth landmark locations (on CelebA). True positives are considered for keypoints within $10\%$ of inter-ocular distance to a landmark location. We observe that generic detectors produce a large number of keypoints that overlap with manually annotated landmarks. }
\label{fig:detectors}
\end{figure}

\begin{table}[!t]
\renewcommand{\arraystretch}{1.3}
\caption{Evaluation of landmarks learned from the first stage of our approach on LS3D~\cite{Bulat2017HowFA} under various number of training clusters. $M$. All models are trained for $K=30$. We see that $M \gg K$ results in better performance as it allows appearance and viewpoint variations of the same landmark to be captured by several clusters }\label{table:clustersAbl}
\centering
\begin{tabular}{lcc}
\toprule
 \# of clusters  & Forward-NME & Backward-NME\\
\toprule
     $M=30$  & 10.26 & 9.41\\
     $M=50$  & 7.99 & 6.99\\
     $M=100$  & \textbf{7.95} & 6.55\\
     $M=250$  & 8.58 & 6.26\\
     $M=500$  & 9.53 & \textbf{6.19}\\
  \bottomrule
  \end{tabular}
\end{table}

\begin{table}[!t]
\renewcommand{\arraystretch}{1.1}
\caption{ Ablation study of the proposed negative pair selection strategy (compared to the strategy of \cite{unsupervLandm2020}), combined with either clustering or equivariance training.  Experiment performed in the challenging LS3D~\cite{Bulat2017HowFA} dataset. We report forward-NME error values. }\label{table:equivariance}
\centering
 \begin{tabular}{lll|c}
\toprule
  \# & \textbf{Negative-Pairs} & \textbf{Correspondence}  & \textit{NME}\\
  \midrule
       1 & \textit{different clusters} & \textit{Clustering}   & 11.15\\
     2 & \textit{same image only} & \textit{Equivariance}    & 10.02\\
      3 & \textit{different clusters} &\textit{Equivariance}   & 9.56\\

      4 & \textit{same image only} &\textit{Clustering}  &  \textbf{7.95}\\
  \bottomrule
  \end{tabular}

\end{table}

\begin{table}[!t]
\renewcommand{\arraystretch}{1}
\caption{ Comparison of the first and second stages of our framework in terms of Forward-NME. We also report average number of points detected per image (p.p.e) on each stage. The full landmark detector on the second stage detects one landmark per $K$ channels so p.p.e is 30. }\label{table:firstsecond}
\centering
 \begin{tabular}{cc|ccc|c}
\toprule
  \textbf{Dataset} & \multicolumn{2}{c}{\textbf{p.p.e}} & & \multicolumn{2}{c}{\textbf{NME($\%$)}}\\
  & Stage1 & Stage2 & & Stage1 & Stage2\\
  \midrule
      \textit{CelebA} ($K=30$)  & \textit{25.8} & \textit{30}  & & 3.3 & \textbf{3.2}\\
      \textit{AFLW} ($K=30$)  & \textit{23.4} & \textit{30} & & 8.1 & \textbf{7.4}\\
      \textit{LS3D} ($K=30$)  & \textit{23.5} & \textit{30} & & 7.9 & \textbf{5.2}\\
  \bottomrule
  \end{tabular}

\end{table}

\begin{table}[!t]
\setlength{\tabcolsep}{3pt}
\centering
\begin{tabular}{ccccc|c}
   & \multicolumn{4}{c}{\textit{LS3D-Test}} & \textit{\textit{CelebA-Test}} \\
   \toprule
  \textbf{TrainSet} & $[0^{\circ},30^{\circ}]$ & $[30^{\circ},60^{\circ}]$ & $[60^{\circ},90^{\circ}]$ & total & total\\
\cmidrule{2-6}
 \textit{CelebA} & 5.86 & 6.21 & 9.20 & 6.95 & 3.25\\
 \textit{LS3D} & 4.42 & 5.07 & 6.51 & 5.26 & 3.33\\
  \bottomrule
  \end{tabular}
  \caption{Cross-Dataset evaluation. We report the forward-NME on the 
 test partitions of CelebA and LS3D. For LS3D-Test (LS3D-Balanced), error is shown across poses (measured in buckets of different yaw angles). }
  \label{table:generalisation}
\end{table}

\begin{table}[!t]
\caption{ Experiments on the effect of flipping as a training augmentation and at test time. Results are given for both stages of our approach in terms of Forward-NME.}\label{table:flipping}
\centering
 \begin{tabular}{ccccc}
\toprule
  Dataset  & Flip\textit{(Train)} & Flip\textit{(Test)} & \textbf{Stage1} &  \textbf{Stage2}\\
   \midrule
   \multirow{3}{*}{ \textit{CelebA}} & \xmark & \xmark & 3.88 & 3.42\\
    & \cmark & \xmark & 3.32 & 3.40\\
    & \cmark & \cmark & 3.32 & 3.25\\
   
   \midrule
    \multirow{3}{*}{ \textit{LS3D}} & \xmark & \xmark & 8.69 & 5.81\\
    & \cmark & \xmark & 7.95 & 5.45\\
    & \cmark & \cmark & 7.95 & 5.26\\

  \bottomrule
  \end{tabular}

\end{table}

\textbf{Landmarks captured as keypoints}: To further evaluate the quality of different keypoint initialisations, we measure to which extent each detector provides keypoints that are consistently close to a manually annotated landmark. To do so, we compute the precision of each of the detectors, measured as the percentage of keypoints that lie  within radius equal to $10\%$ of inter-ocular distance to a groundtruth point. Fig. \ref{fig:detectors} shows some visual examples of keypoints that overlap with manually annotated landmarks (red), as well as the computed precision. These results align with those in Table~\ref{table:detectorAbl}, showing that SuperPoint is a better choice to populate the training set.

\subsection{On the training design}

\textbf{Impact of number of clusters}: We investigate the effect on the number of clusters in the training of our proposed approach. The results shown in Table~\ref{table:clustersAbl} indicate that the best performance is attained for a larger number of training clusters. This over-segmentation of feature space is required for optimal clustering assignment as it allows for multiple clusters that capture different appearance variations of the same landmark, enabling the discovery of more stable landmarks (as demonstrated by smaller values of the backward error in Table ~\ref{table:clustersAbl}). On the other extreme, for very big $M$ values, the same underlying landmark is tracked by several clusters, each containing only very similar features. This hinders our method's ability to learn representations robust to viewpoint or appearance variations, and more diverse landmarks get filtered out (leading to an increase in Forward-NME). Note that our method essentially equates to equivariance training in extreme cases where $M$ is equal to the number of detected keypoints (each cluster contains only one feature).

\textbf{Negative-Pair Selection}: We evaluate the proposed negative pair selection strategy (referred to as \textit{same image only}), compared to that of ~\cite{unsupervLandm2020} (referred as \textit{different cluster}) where negative pairs were selected as keypoints with different clustering assignments. We also evaluate the effect of learning from unpaired images (enabled by correspondence recovery) compared to training on synthesised views of the same underlying image (equivariance training). Note that the experiments that use equivariance still utilise deep clustering  (constraint the detector in detecting at most $K$ landmarks and filtering out noisy keypoints). Results can be seen in Table \ref{table:equivariance}.

We observe that our improved negative pair selection strategy is the best performing method when correspondence is recovered through clustering (\textit{line 4}). The \textit{different cluster} strategy separates features to $M$ clusters (\textit{line 1}) and results in high error when is not combined with an additional merging step (as in \cite{unsupervLandm2020}). Also, our negative pair selection strategy is only beneficial when correspondence is recovered through clustering (not with equivariance). This is expected since, with equivariance training, point correspondences are known, and inaccurate negative pairs (similar to the ones shown in Fig. \ref{fig:pairs}) do not emerge. As a result negative pairs from \textit{different cluster} are more informative and result in lower error values (\textit{line 3} vs. \textit{line 2}). We also present in Fig.~\ref{fig:tsne} the t-SNE~\cite{Maaten2008VisualizingDU} representations of features returned by SuperPoint (left), by ~\cite{unsupervLandm2020} (center), and our proposed method (right). Our method produces features that are clearly distinctive for each landmark, making the correspondence recovery effective.

\begin{figure*}
\begin{subfigure}{\linewidth}
\centering
\includegraphics[trim=125 0 30 10,width=\linewidth]{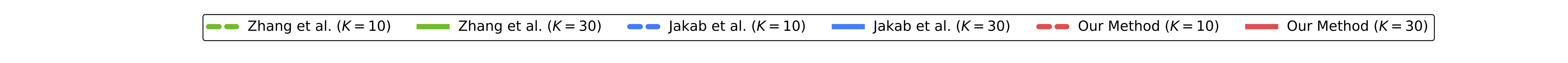}
\end{subfigure}
\begin{subfigure}{.48\linewidth}
\begin{subfigure}{\linewidth}
\begin{subfigure}[b]{.48\linewidth}
\includegraphics[trim=0 0 0 0,clip,width=\linewidth]{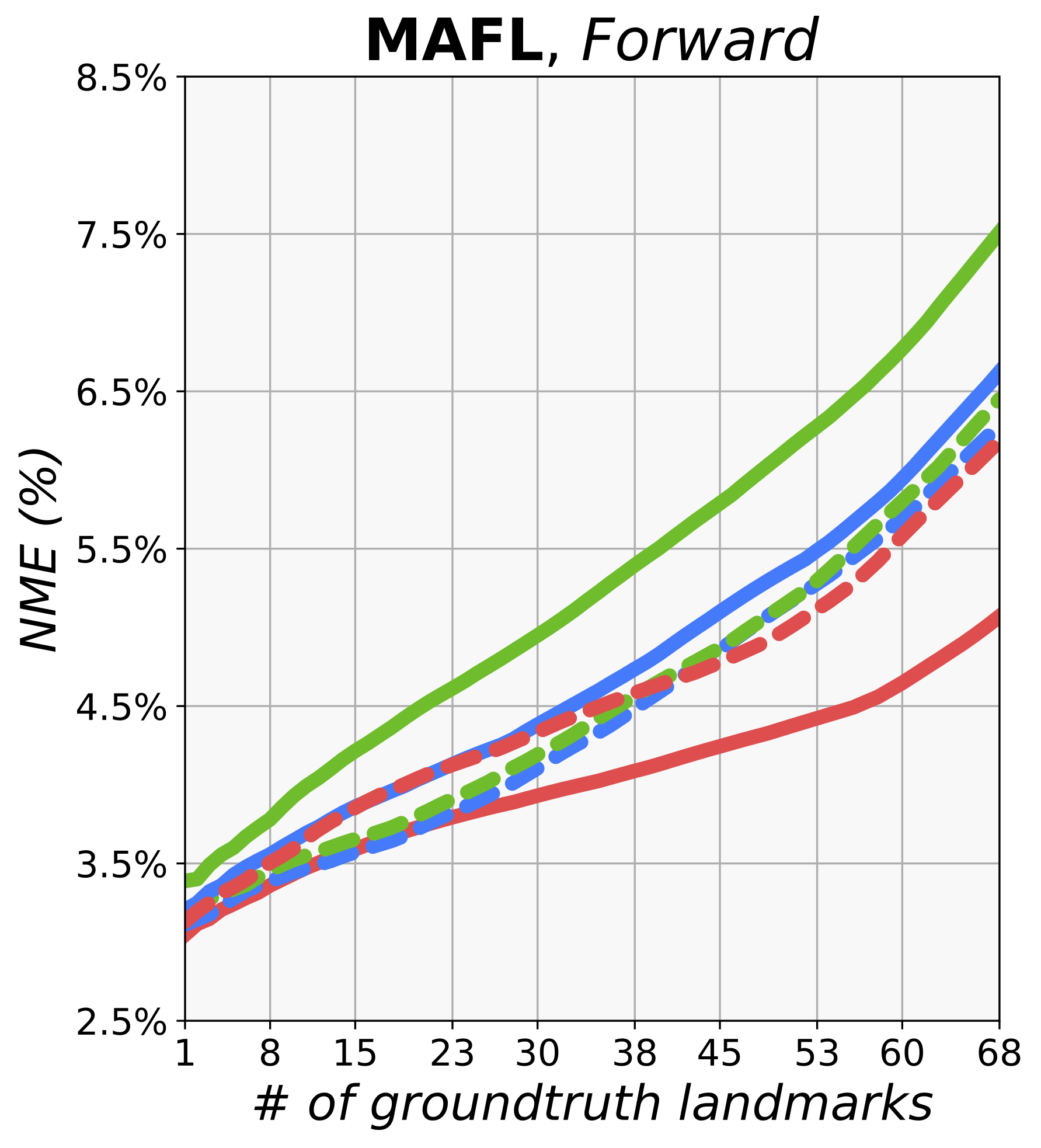}
\end{subfigure}
\begin{subfigure}[b]{.48\linewidth}
\includegraphics[trim=0 0 0 0,clip,width=\linewidth]{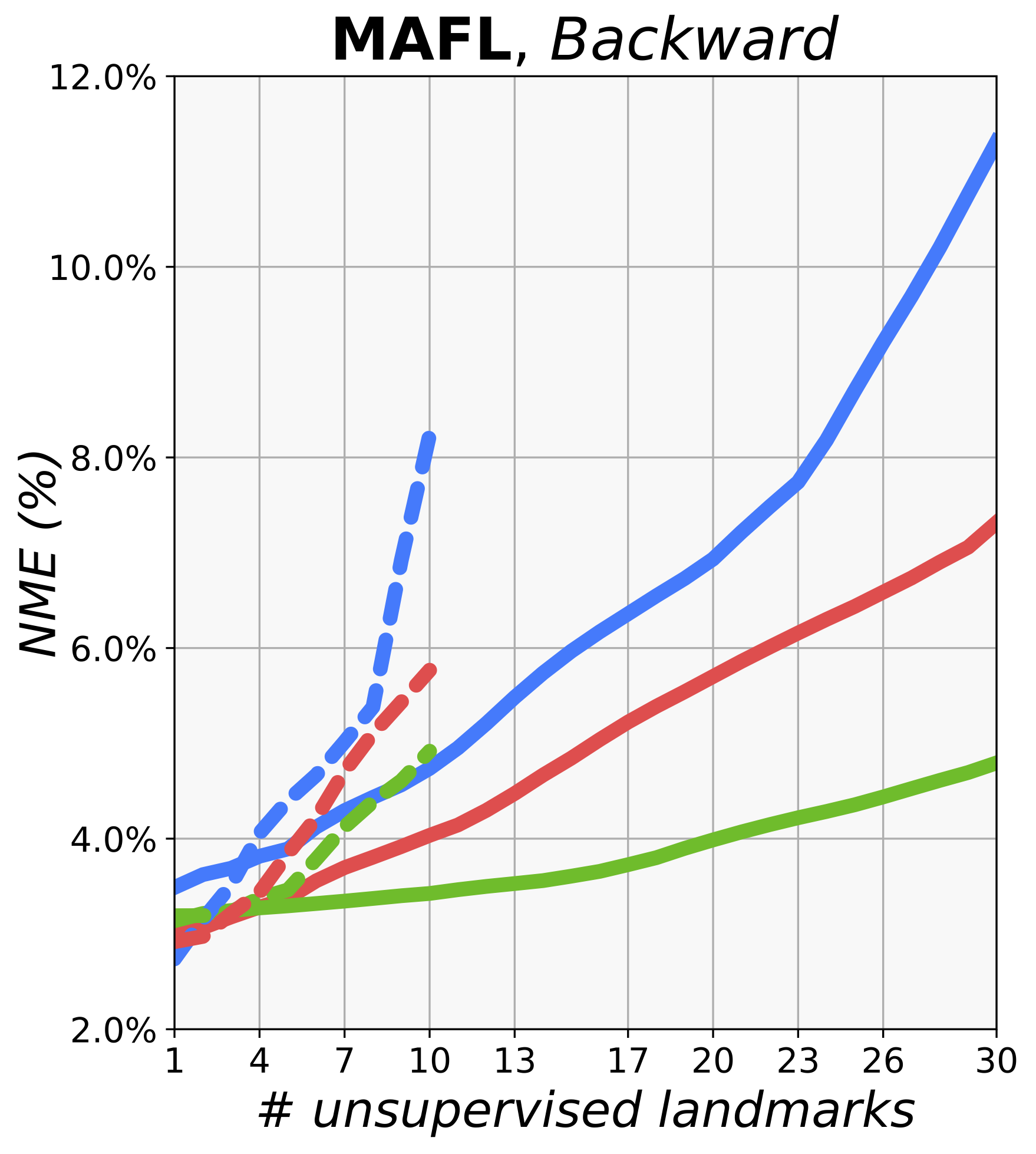}
\end{subfigure}
\end{subfigure}
\end{subfigure}
\begin{subfigure}{.48\linewidth}
\begin{subfigure}{\linewidth}
\begin{subfigure}[b]{.48\linewidth}
\includegraphics[trim=0 0 0 0,clip,width=\linewidth]{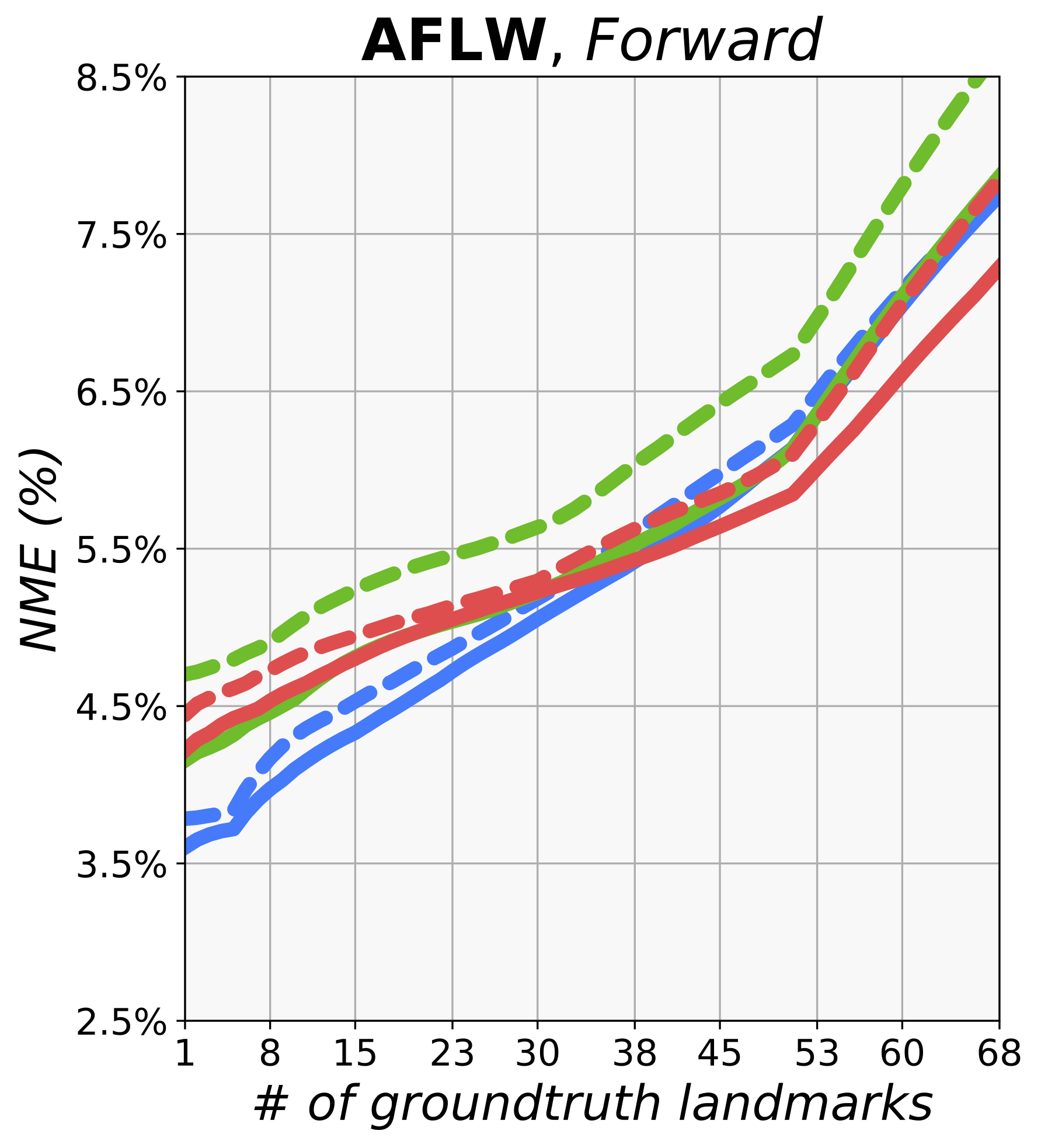}
\end{subfigure}
\begin{subfigure}[b]{.48\linewidth}
\includegraphics[trim=0 0 0 0,clip,width=\linewidth]{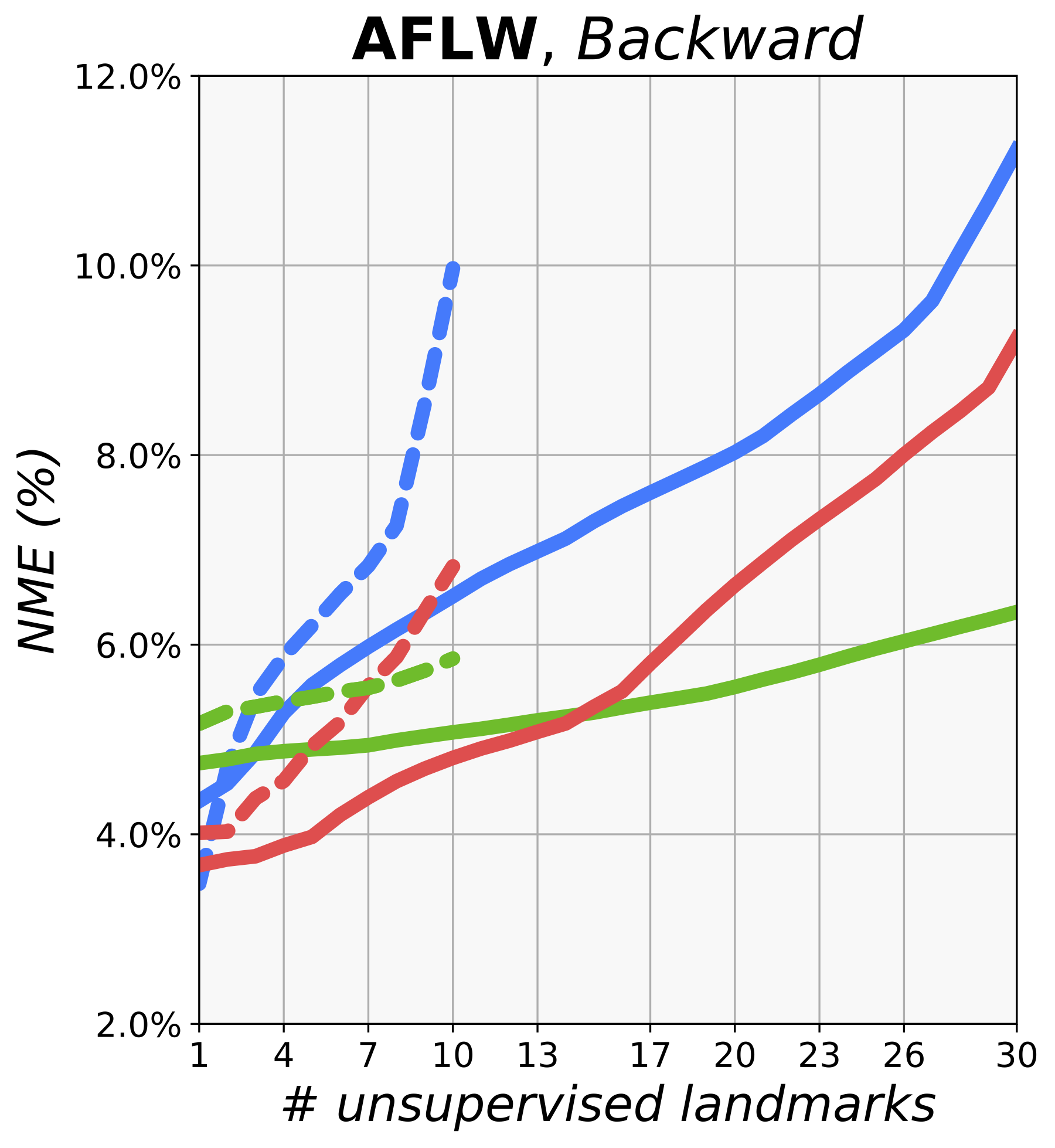}
\end{subfigure}
\end{subfigure}
\end{subfigure}
\vskip 0.2in
\begin{subtable}{.48\linewidth}
\centering
\setlength\tabcolsep{1.5pt}
\begin{tabular}{lcc}
 \textbf{Method}  & MAFL & AFLW\\
 \midrule
  Lorenz~\cite{Lorenz2019UnsupervisedPD} \textit{(K=10)}& 3.24 & -\\
  Shu~\cite{Shu2018DeformingAU}  & 5.45 & - \\
  Jakab et al.\cite{jakab2018unsupervised} \textit{(K=10)} & 3.19 & 6.86\\
  Zhang et al.~\cite{Zhang2018UnsupervisedDO} \textit{(K=10)} & 3.46 & 7.01\\
  Sanchez~\cite{NIPS2019_9505} \textit{(K=10)} & 3.99 & 6.69\\
  Sahasrabudhe~\cite{Sahasrabudhe2019LiftingAU} & 6.01 & -\\
    Mallis~\cite{unsupervLandm2020} & 4.12 & 7.37\\
  \midrule
Ours \textit{(K=10)} & 3.83 & 7.18\\
  \bottomrule
  \end{tabular}
\end{subtable}
\begin{subfigure}{.48\linewidth}
\begin{subfigure}{\linewidth}
\begin{subfigure}[b]{.48\linewidth}
\includegraphics[trim=0 0 0 0,clip,width=\linewidth]{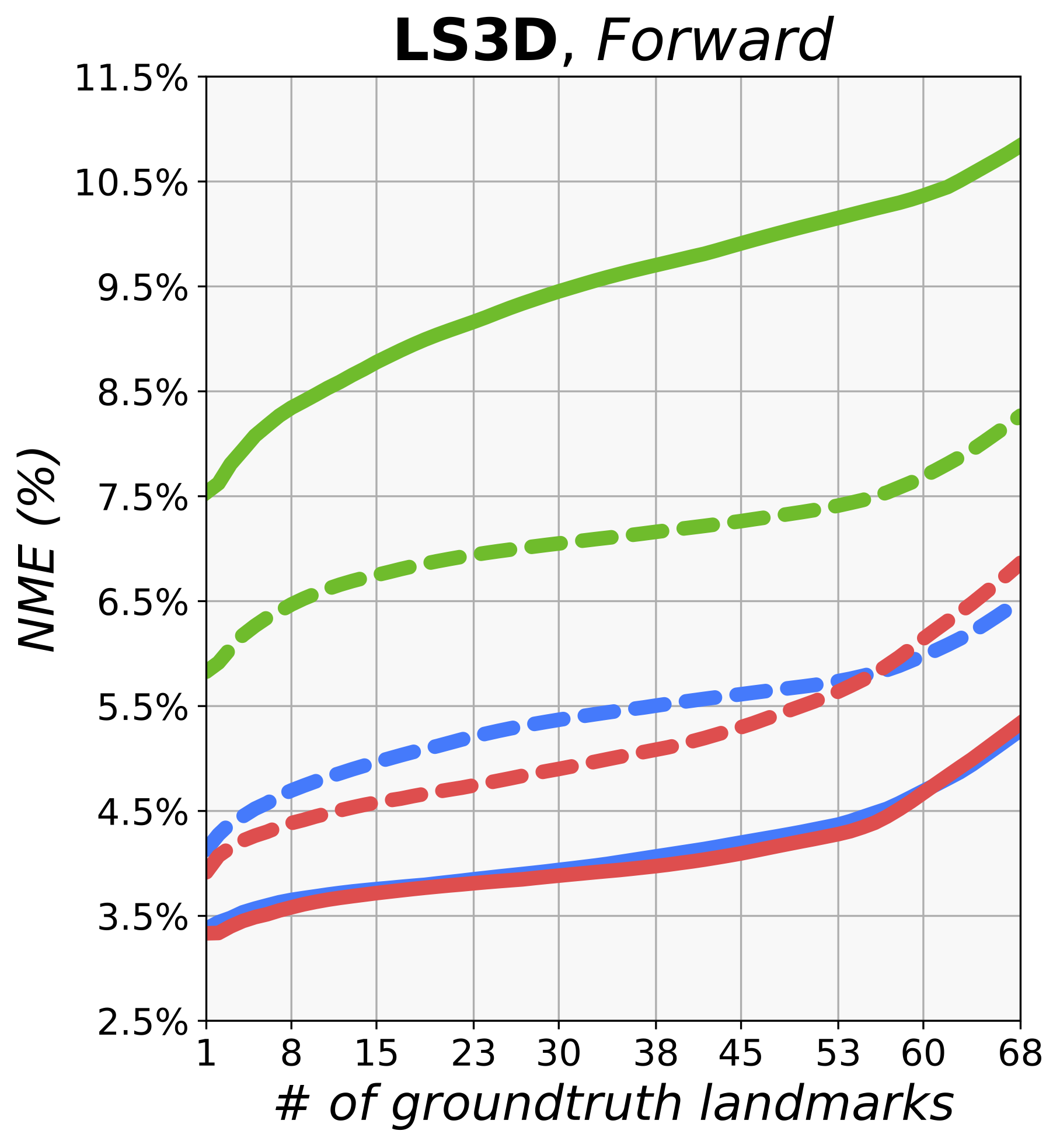}
\end{subfigure}
\begin{subfigure}[b]{.48\linewidth}
\includegraphics[trim=0 0 0 0,clip,width=\linewidth]{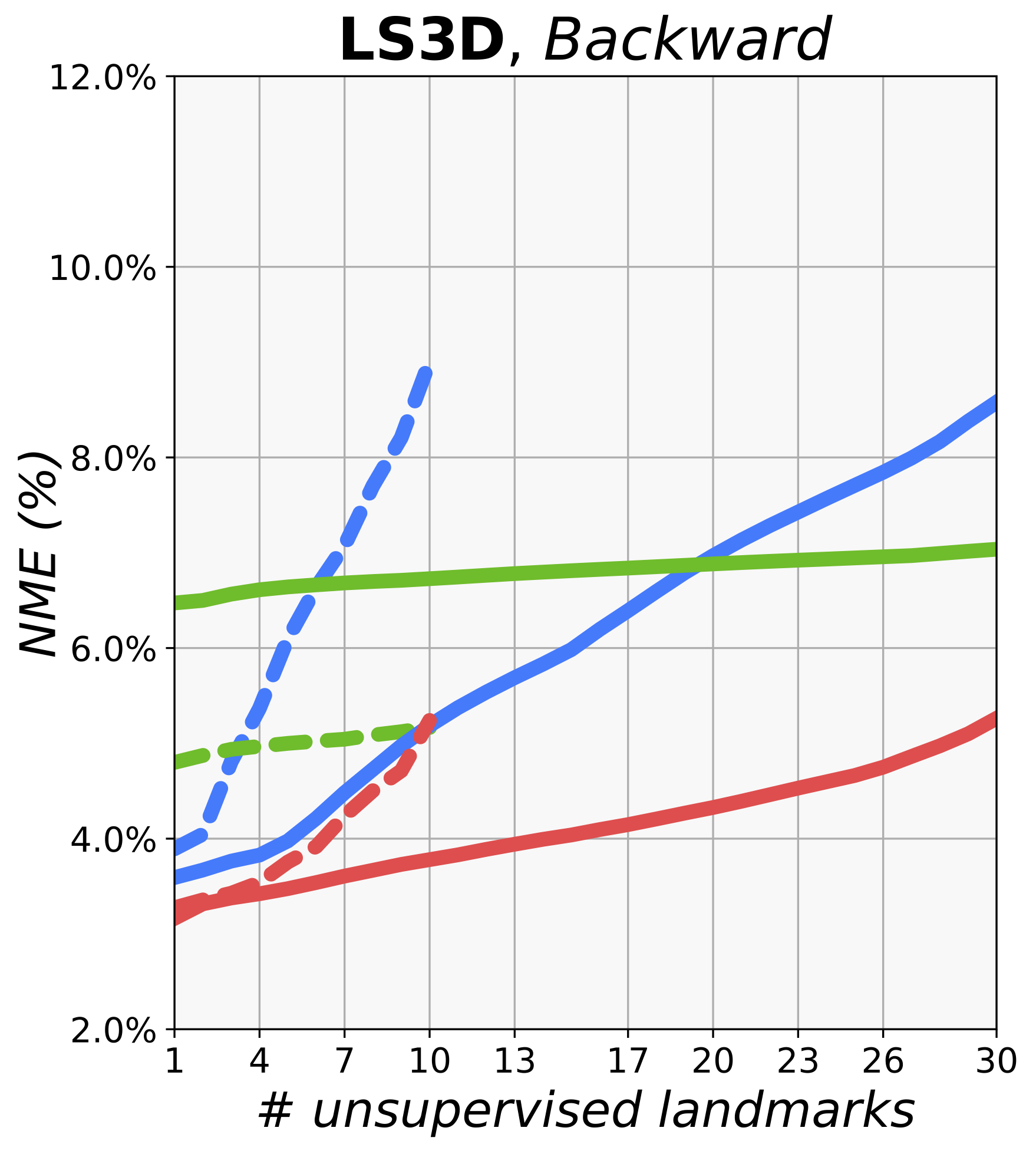}
\end{subfigure}
\end{subfigure}
\end{subfigure}
\caption{Evaluation on facial datasets. \textit{\textbf{(Table):}} Standard comparison on MAFL and AFLW, in terms of forward error. The results of other methods are taken directly from the papers (for the case where all MAFL training images are used to train the regressor and the error is measured w.r.t. to $5$ annotated points). \textit{\textbf{(Figures):}} CED curves for forward and backward errors. We compare our method with~\cite{jakab2018unsupervised,Zhang2018UnsupervisedDO} (for $K=10,30$). Where possible, we used pre-trained models, otherwise we re-trained these methods using the publicly available code. A set of $300$ training images is used to train the regressors. Error is measured w.r.t. the $68$-landmark configuration typically used in face alignment.   }\label{ref:facialresults}
\end{figure*}

\begin{figure*}
\centering
\includegraphics[width=\linewidth]{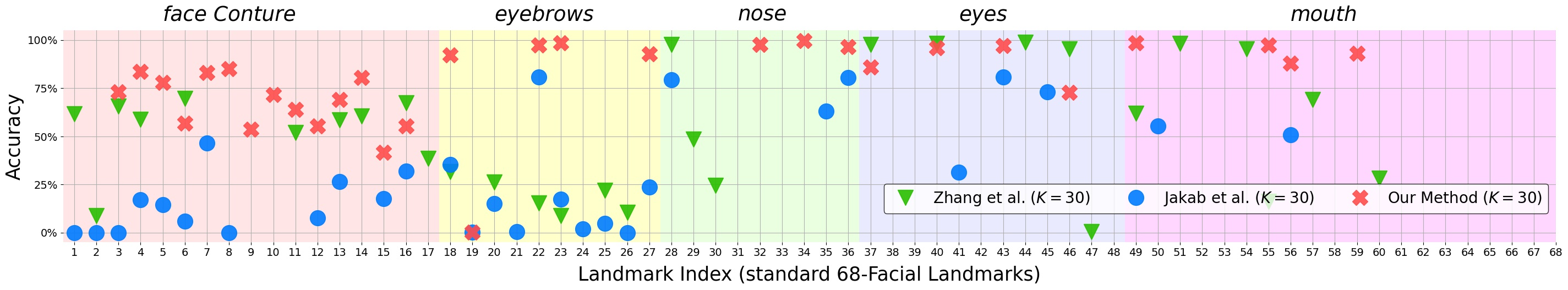}
\caption{Evaluation of the ability of raw unsupervised landmarks to capture supervised landmark locations on CelebA. Each unsupervised landmark is mapped to the best corresponding supervised landmark using the Hungarian Algorithm. Then accuracy is calculated for a distance threshold of $0.2 \cdot d_{iod}$ to a landmark location, where $d_{iod}$ is the interocular distance. Accuracy is shown for each of the 68-facial landmarks sorted by ascending order of index. Different landmark areas are highlighted with different colours (1-17 are facial contour landmarks, 18-27 are landmarks tracking the eyebrows, etc.) }\label{ref:facialaccuracy}
\end{figure*}

\textbf{Impact of Stage 2}: For the first stage of our method, a set of points are detected per image for which correspondence is recovered through clustering. In the second stage, these points and correspondences are used to train a landmark detector with $K$ output channels. The number of detected points per image on the first stage is $\leq K$ since there is no guarantee that each would appear in each image. On the contrary, our full landmark detector (output of the second stage) learns $K$ unsupervised landmarks (one per output heatmap). In Table \ref{table:firstsecond} we compare performance of the first vs second stage in terms of forward NME while also report the average number of points detected per image. We observe that the full landmark detector recovers the missing clusters in the second stage, resulting in lower error values. Performance increase is most notable on LS3D, where occlusion is extended due to large jaw angles.

\textbf{Generalisation}: To further analyse the robustness of our proposed approach to in-plane rotations as well as to domain shift, we conduct a cross-dataset evaluation (Table~\ref{table:generalisation}). In particular, for $K=30$, we evaluate models trained on CelebA and LS3D on the corresponding test partitions (LS3D-Balanced and MAFL). In Table~\ref{table:generalisation}, we break down the results for LS3D-Balanced in different yaw angle ranges (to allow for a better comparison between the same-dataset results shown in Table \ref{table:firstsecond} and those given by this cross-dataset evaluation). We can observe that due to the lack of in-plane rotations on CelebA, the model tends to produce high error values for larger poses in LS3D-Test. On the contrary, the model trained on LS3D can maintain its robustness on the CelebA-Test partition, given that it is composed of mostly frontal faces. We can also observe that the improvement of the LS3D model compared to that trained on CelebA, is quite significant on the most difficult partition of the LS3D-Test, illustrating the need of having a diverse set of images describing the geometry of the target object.

\textbf{Flipping}: Finally, we conduct an ablation study on the proposed flipping augmentation strategy. Results for both CelebA and the more challenging LS3D database are given in Table \ref{table:flipping}. We observe that both flipping as a training augmentation and flipping at test time result in a consistent error reduction in terms of Forward-NME.

\begin{figure}
\centering
\includegraphics[width=0.93\linewidth]{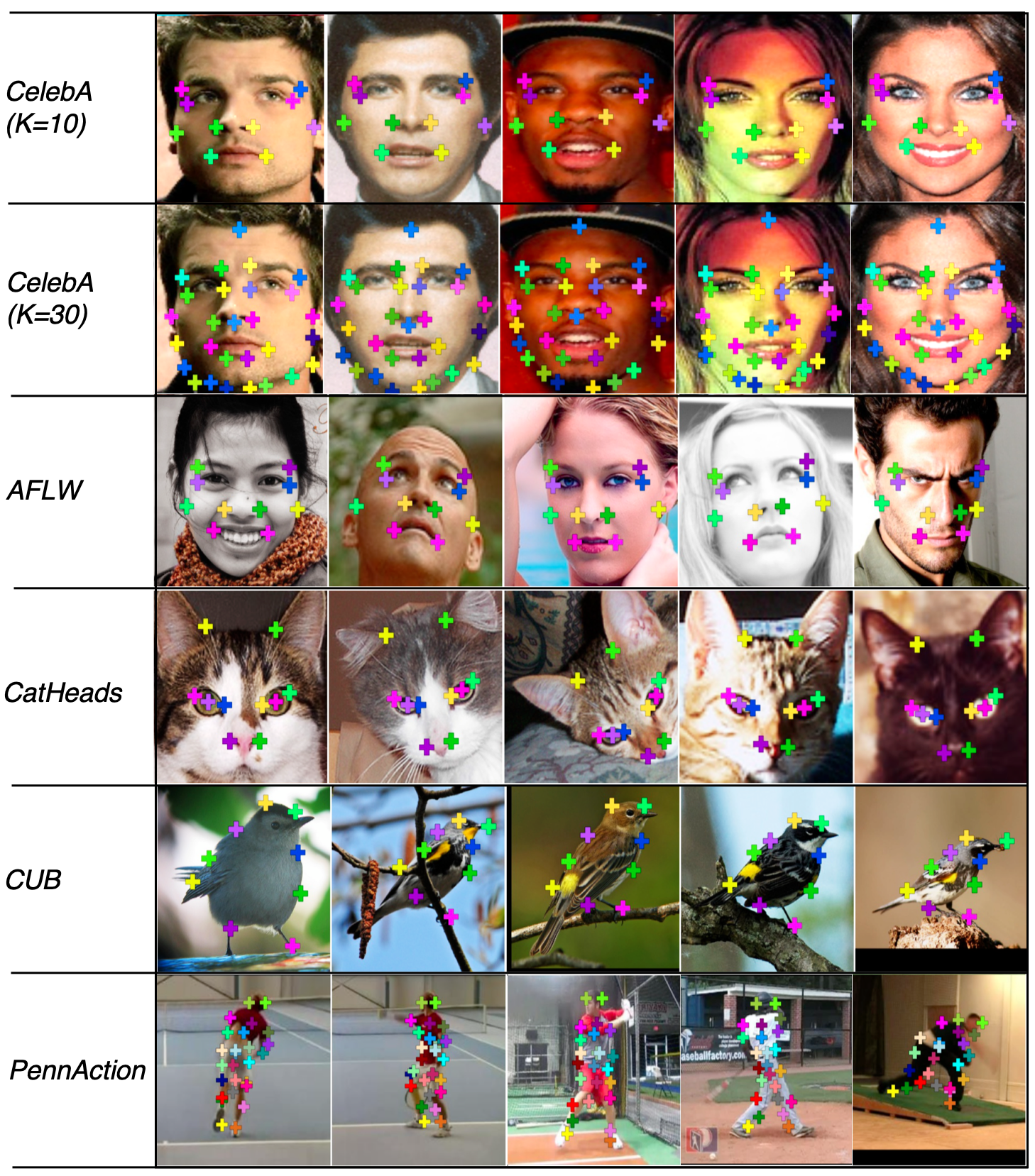}
\caption{Qualitative results of our proposed approach on various object categories. We consistently discover points in key parts corresponding to the eye corners or the contour. Our method assigns a proper landmark index to these points firstly discovered by a keypoint detector and refined through our self-training approach. }\label{ref:visualres}
\end{figure}

\begin{figure}
\centering
\includegraphics[width=0.93\linewidth]{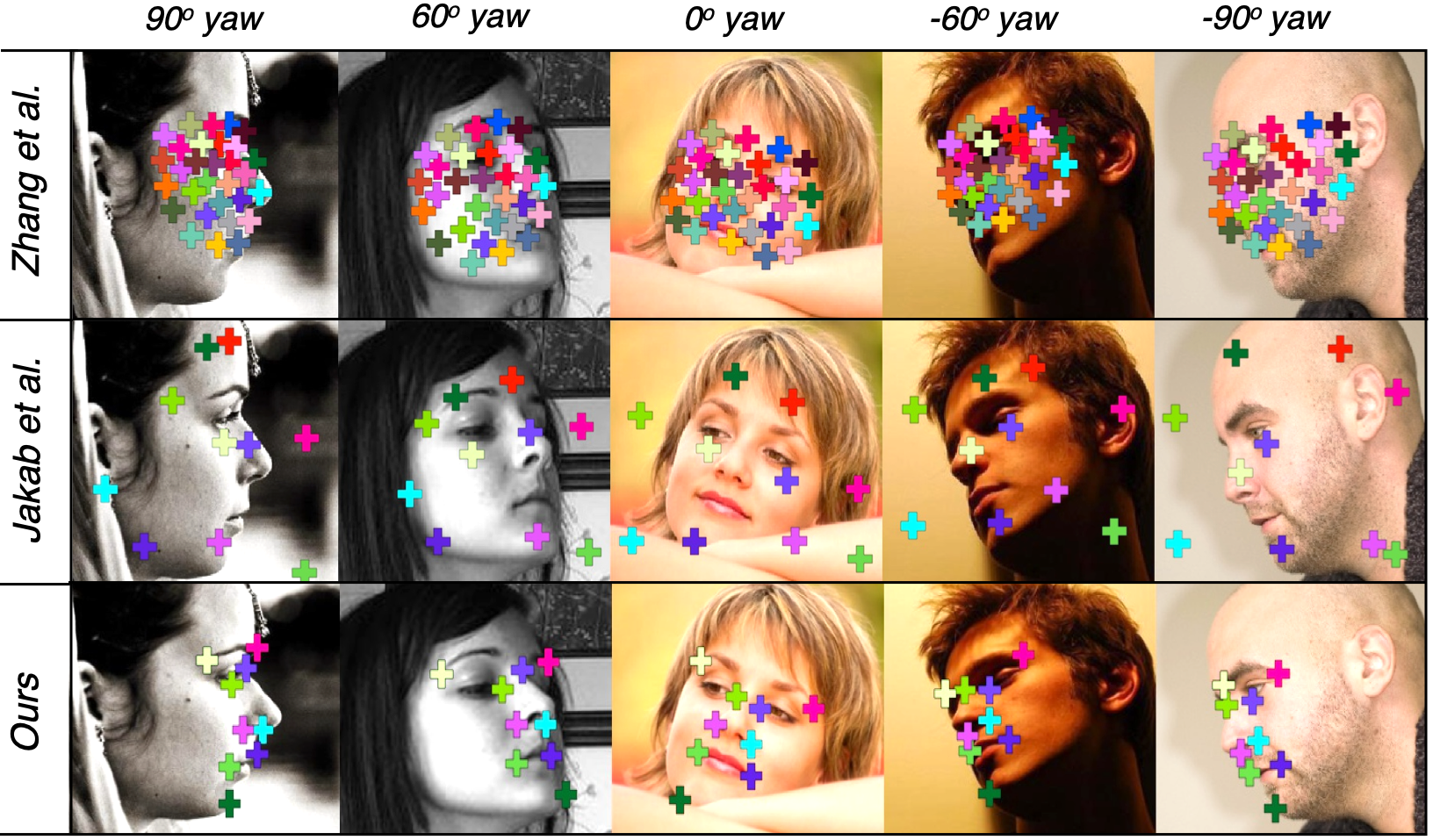}
\caption{Visual examples of landmarks discovered by \cite{Zhang2018UnsupervisedDO} (top row), \cite{jakab2018unsupervised} (middle row), and our method (bottom), on LS3D, across a variety of poses. While \cite{Zhang2018UnsupervisedDO,jakab2018unsupervised} fail to model large viewpoint changes, our method benefits from having descriptors that can model the same semantic landmark. }\label{ref:jaw}
\end{figure}

\begin{figure*}[!t]
\begin{subfigure}{\linewidth}
\centering
\includegraphics[width=0.6\linewidth]{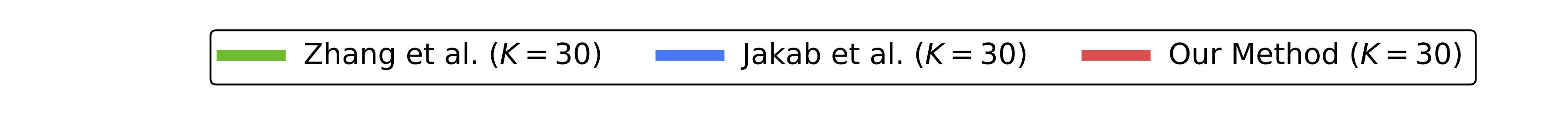}
\end{subfigure}
\begin{subfigure}{.48\linewidth}
\begin{subfigure}{\linewidth}
\begin{subfigure}[b]{.48\linewidth}
\includegraphics[trim=0 0 0 0,clip,width=\linewidth]{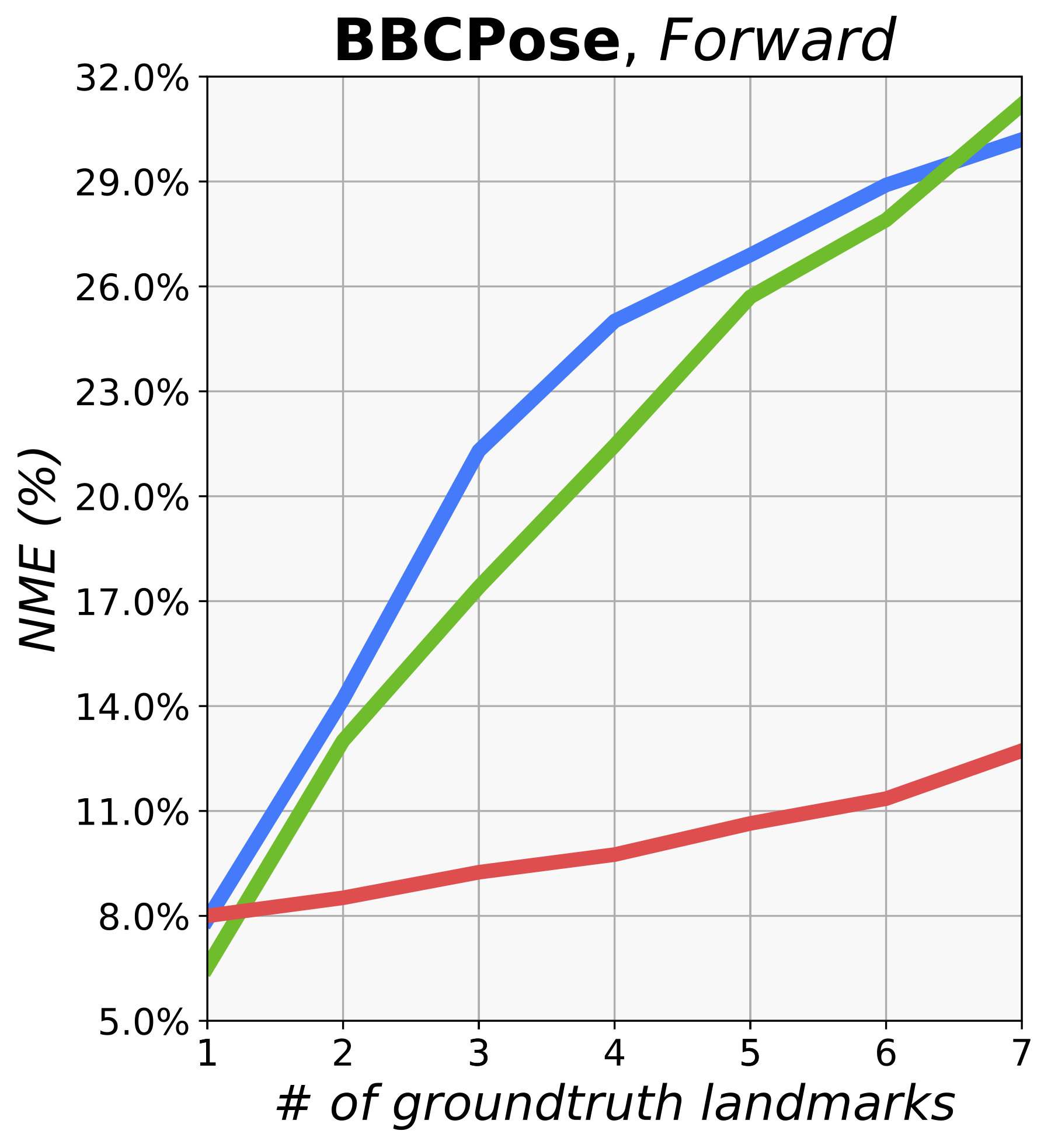}
\end{subfigure}
\begin{subfigure}[b]{.48\linewidth}
\includegraphics[trim=0 0 0 0,clip,width=\linewidth]{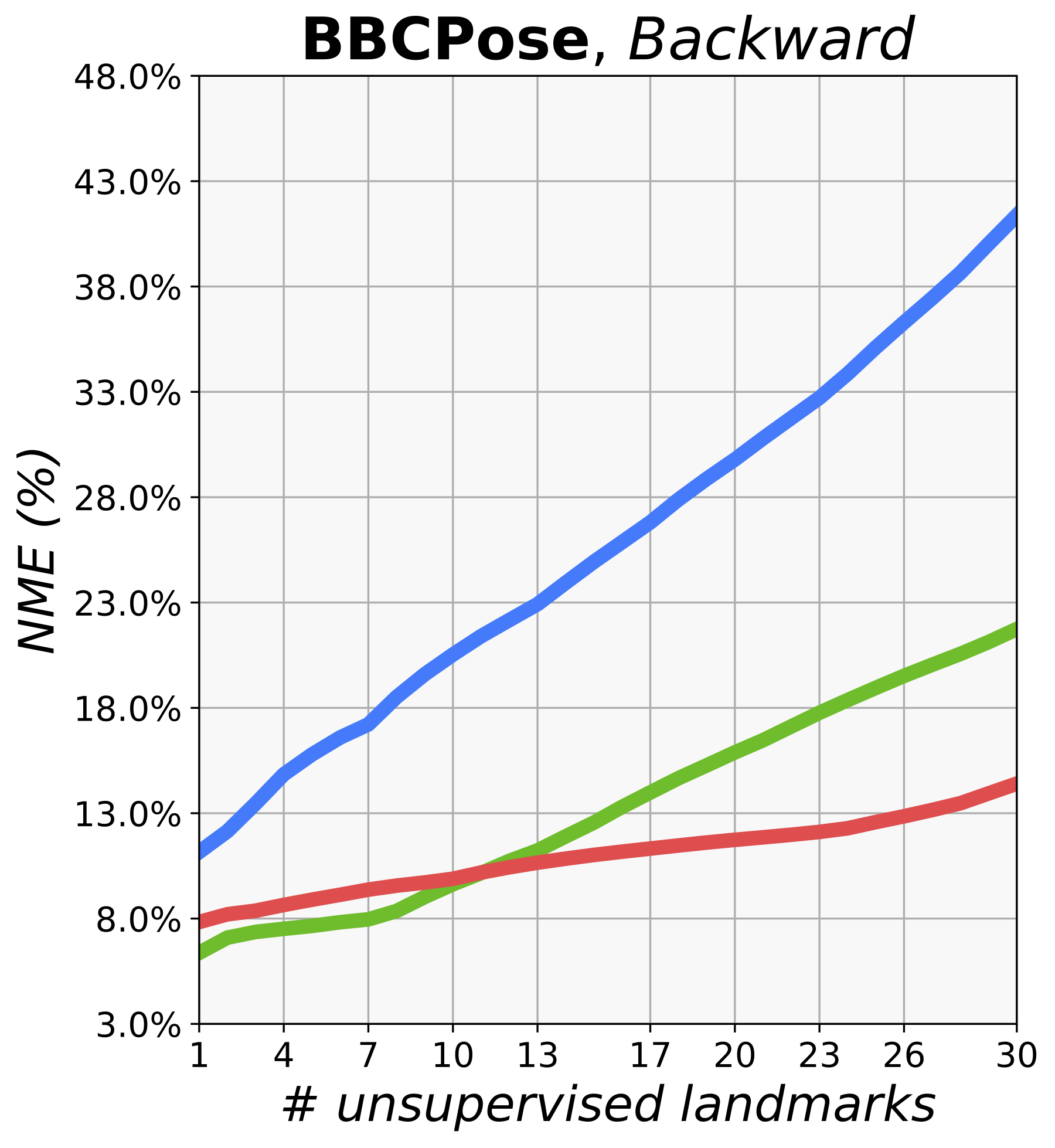}
\end{subfigure}
\end{subfigure}
\end{subfigure}
\begin{subfigure}{.48\linewidth}
\begin{subfigure}{\linewidth}
\begin{subfigure}[b]{.48\linewidth}
\includegraphics[trim=0 0 0 0,clip,width=\linewidth]{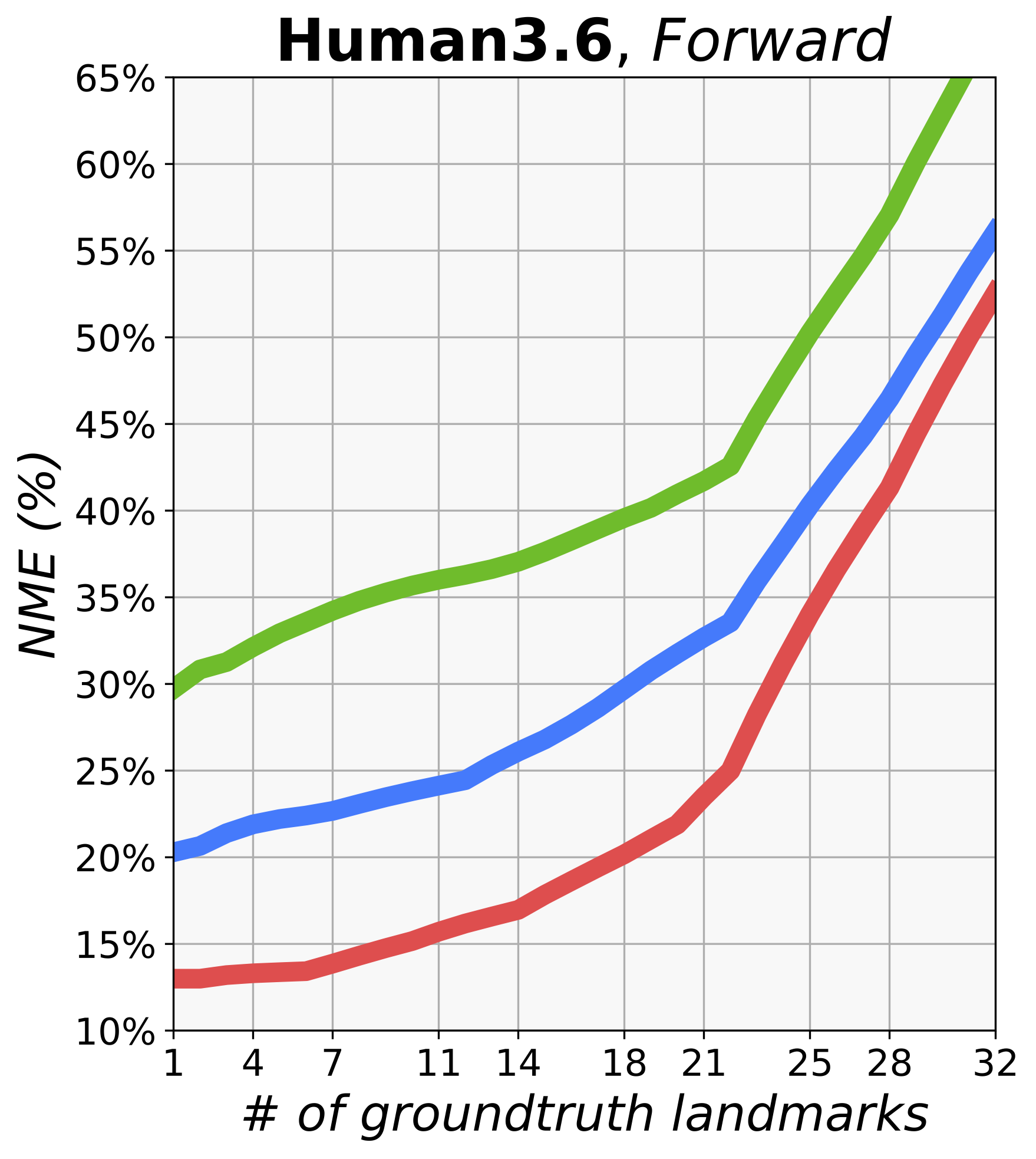}
\end{subfigure}
\begin{subfigure}[b]{.48\linewidth}
\includegraphics[trim=0 0 0 0,clip,width=\linewidth]{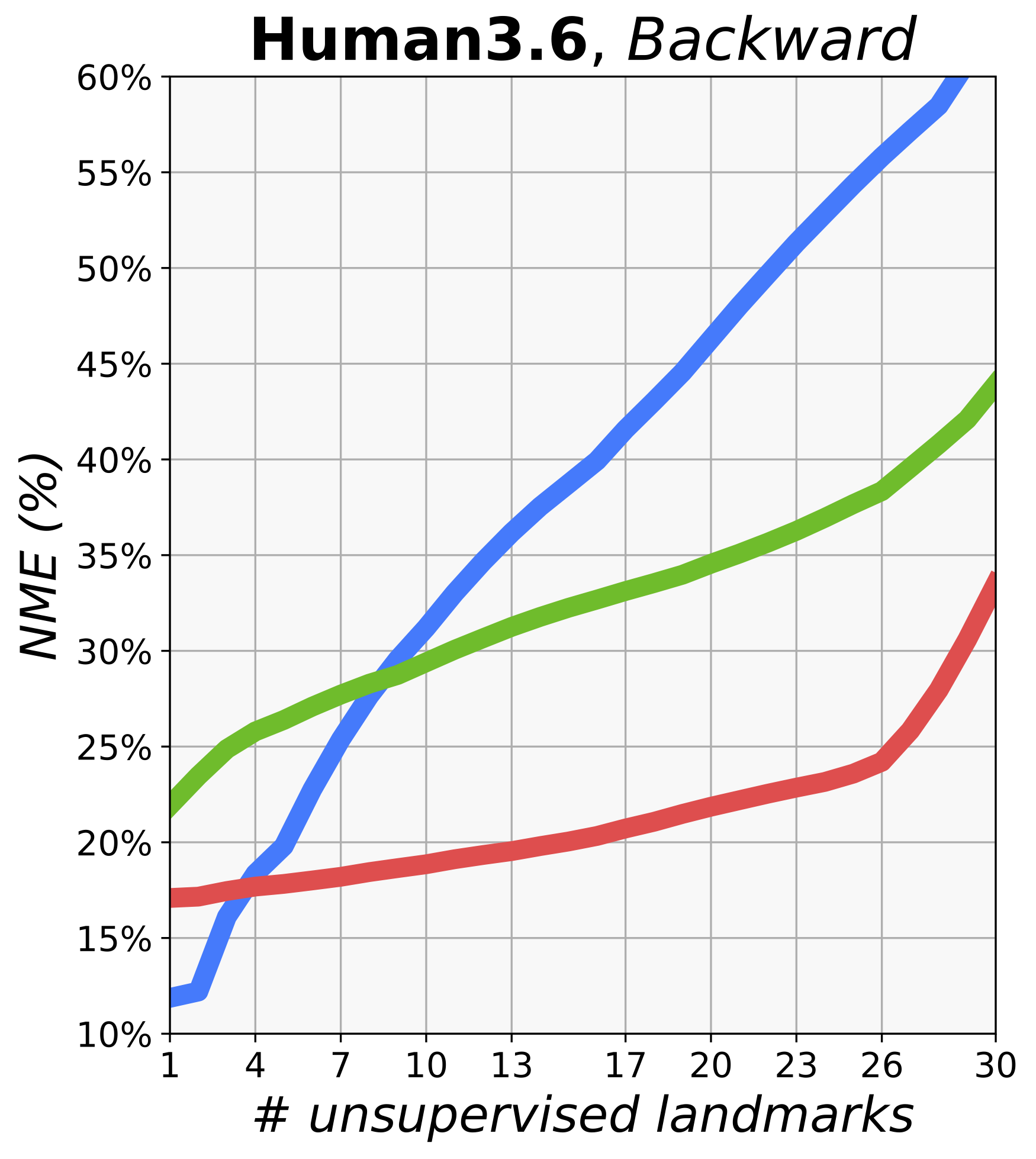}
\end{subfigure}
\end{subfigure}
\end{subfigure}
\caption{Evaluation on BBCPose and Human3.6 datasets. CED curves for the forward and backward errors, computed for a regressor trained with $800$ samples. We compare our method with~\cite{jakab2018unsupervised,Zhang2018UnsupervisedDO} (re-trained using the publicly available code). All methods are trained to discover 30 landmarks. }\label{fig:humanpose}
\end{figure*}

\section{Overall evaluation} \label{sec:sota}
This Section presents the experiments carried out to validate the proposed approach against state-of-the-art alternatives based on equivariance or image generation. We report qualitative results on various datasets in Fig. \ref{ref:visualres}.  

\textbf{Evaluation on facial datasets.} Fig.~\ref{ref:facialresults} shows the results of our method on facial datasets. We report in the \textit{Table} the commonly used forward error w.r.t. the 5 ground-truth facial landmarks. For the cumulative curves, the error is calculated w.r.t. 68-standard facial landmarks. As discussed in~\cite{NIPS2019_9505}, for a method to work well, both forward and backward errors should be small. From our results on all datasets (\textit{Figures}), we can see that overall our method provides the best results in terms of meeting both requirements. Notably, our method delivers state-of-the-art results for the challenging LS3D dataset, which contains large pose changes.

We also find that our approach surpasses other methods when evaluation is performed w.r.t all 68-facial landmarks (compared to standard 5 landmark evaluation on MAFL and AFLW presented in Fig.~\ref{ref:facialresults} (\textit{Table}) where we maintain competitive performance). One reason is that 5 facial landmarks include points in uniform areas and not repeatable edges or corners (centre of the eye, centre of the nose) that are not commonly tracked by generic keypoint detectors. On the contrary, our method is better suited to track the 68 commonly used facial landmarks. To further demonstrate that, we evaluate how accurately raw unsupervised landmarks track supervised landmark locations in Fig. \ref{ref:facialaccuracy}. Each of the 68-facial landmarks is matched to the best corresponding unsupervised landmarks ($K=30$ is used for all methods) through the Hungarian algorithm. We observe that most of our detected unsupervised landmarks track actual semantic object locations with high accuracy. In contrast, landmarks detected by \cite{jakab2018unsupervised,Zhang2018UnsupervisedDO} are mostly uniformly spread over the objects' surface (to ensure stronger image generation/reconstruction) and do not tend to track manually annotated landmark locations.

\begin{table}
\centering
\setlength\tabcolsep{10pt}
\begin{tabular}{cccc}

\multicolumn{4}{c}{\textbf{CatHeads Forward-NME ($\%$)}}\\
\toprule
  Thewlis\cite{Thewlis2017UnsupervisedLO} & Zhang\cite{Zhang2018UnsupervisedDO} & Lorenz\cite{Lorenz2019UnsupervisedPD} & Ours\\ 

   26.94 & 14.84 & 9.30 & 9.31\\

  \bottomrule
  \end{tabular}
\caption{Forward-NME on the CatHeads dataset \cite{Zhang2008CatHD}. All methods detect $K=20$ unsupervised landmarks. Results for other methods are taken directly from the papers. Same as other methods, we regress 7 of the 9 annotated landmarks for this experiment (excluding landmarks on the ears). }\label{table:cats}
\end{table}

Evaluation in terms of Forward-NME for the CatsHead dataset is shown in Table. \ref{table:cats}. Our method reaches a similar error value as the best performing method of \cite{Lorenz2019UnsupervisedPD}.
In addition, a set of qualitative examples is shown in Fig.~\ref{ref:jaw} for the challenging LS3D data. We observe that landmarks produced by \cite{jakab2018unsupervised,Zhang2018UnsupervisedDO} are not stable under 3D rotations and fail to capture large pose variations. 

\begin{figure*}[!t]
\centering
\includegraphics[width=\linewidth]{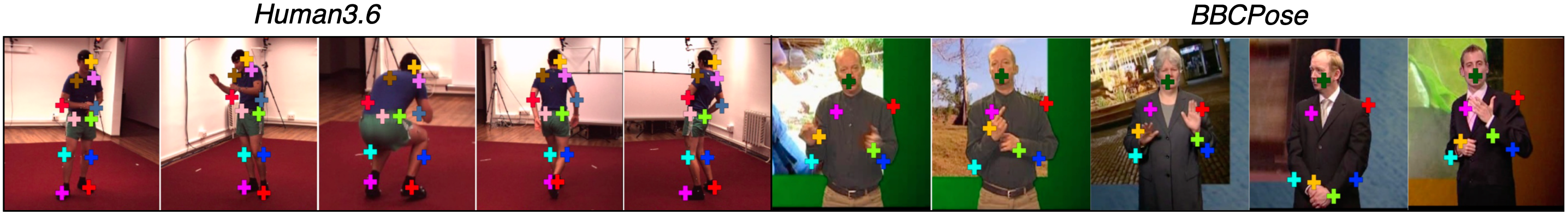}
\caption{Examples on Human3.6 and BBCPose databases. We show the unsupervised landmarks that maximally corresponding to the provided ground-truth (selected through the Hungarian Algorithm).  }\label{ref:maximal}
\end{figure*}

\begin{table}
\centering
\setlength\tabcolsep{1.5pt}
\begin{tabular}{lccccc}
\multicolumn{6}{c}{\textbf{BBCPose Regressed Landmark Accuracy ($\%$)}}\\
\toprule
 \textbf{Method} & Head & Shldrs & Elbws & Hands  & Avg\\
   \midrule
 \multicolumn{6}{c}{\textit{Supervised}}\\
  \noalign{\vskip 1mm}    
   Yang~\cite{Yang2011ArticulatedPE} & 63.40 & 53.70 & 49.20 & 46.10 & 51.63\\
Pfister~\cite{Pfister2014DeepCN} & 74.90 & 53.05 & 46.00 & 71.40 & 59.40\\
Chen~\cite{Chen2014ArticulatedPE} & 65.90 & 47.90 & 66.50 & 76.80 & 64.10\\
 Charles~\cite{Charles2013DomainAF} & 95.40 & 72.95 & 68.70 & 90.30 & 79.90\\
 Pfister~\cite{Pfister2015FlowingCF} & 98.00 & 88.45 & 77.10 & 93.50 & 88.01\\
  \noalign{\vskip 2mm}    
 \multicolumn{6}{c}{\textit{Unsupervised}}\\
\noalign{\vskip 1mm}  
 Jakab~\cite{jakab2018unsupervised}(selfsup) & 81.01 & 49.05 & 53.05 & 70.10 & 60.79\\
 Jakab~\cite{jakab2018unsupervised} & 76.10 & 56.50 & 70.70 & 74.30 & 68.44\\
 Lorenz~\cite{Lorenz2019UnsupervisedPD} & - & - & - & - & 74.50\\
  \midrule
 Ours & 97.89 & 49.65 & 71.26 & 84.90 & 75.93\\
  \bottomrule
  \end{tabular}
  \caption{ Accuracy of regressed landmarks on BBCPose measured as $\%$-age of points within $d=6px$ from the ground-truth for a resolution of $128px$. Results for other methods taken directly from the papers. All unsupervised methods in this experiment utilise temporal information.}\label{table:bbcaccuracy}
\end{table}

\textbf{Evaluation on human pose datasets} Evaluation of our method on the BBCPose and Human3.6M datasets is shown in Fig. \ref{fig:humanpose}. Note that in this experiment, all methods are trained without temporal supervision. For both datasets, our approach demonstrates significantly lower error values. As it can be seen from the forward error in Human3.6M, all three methods experience a sharp error increase when more than $22$ landmarks are considered. We attribute this higher error to the fact that the hands are not captured by any method. Table \ref{table:bbcaccuracy} we measure the accuracy of regressed landmarks on the BBCPose database. For this experiment temporal supervision is available for all unsupervised methods. Even though this enables other approaches to achieve higher accuracy, our model outperforms all other methods. Fig.~\ref{ref:maximal} shows some examples of discovered landmarks that maximally correspond to ground-truth points.

\begin{table}
\centering
\setlength\tabcolsep{1.5pt}
\begin{tabular}{lccccccc}
\multicolumn{8}{c}{\textbf{PennAction Raw Landmark Accuracy ($\%$)}}\\
\toprule
 \textbf{Method}  & Head & Shldrs & Elbws & Hands & Waist & Knees & Legs \\ 
  \midrule
  Jakab~\cite{jakab2018unsupervised}  & 6.36 & 9.23 & 7.85 & 0.59 & 22.27 & 17.85 & 6.48  \\
  Ours & 74.27 & 57.91 & 33.00 & 8.36 & 64.81 & 69.54 & 75.84    \\
  \bottomrule
  \end{tabular}
\caption{Accuracy of raw discovered landmarks that correspond maximally to each ground-truth point measured as $\%$-age of points within $d=6px$ from the ground-truth (image resolution of $128px$).}  \label{pennraw}
\end{table}

\begin{table}
\centering
\setlength\tabcolsep{1.5pt}
\begin{tabular}{lccccccc}
\multicolumn{8}{c}{\textbf{Human3.6 Raw Landmark Evaluation}}\\
  \toprule
    \multicolumn{8}{c}{\textbf{Accuracy}($\%$)}\\
  \midrule
 \textbf{Method}  & Head & Shldrs & Elbws & Waist & Knees & Legs & Avg\\  
 \midrule
   Zhang~\cite{Zhang2018UnsupervisedDO}  & 20.9 & 53.1 & 51.0 & 43.7 & 85.6 & 2.0 & 42.7 \\
  Jakab~\cite{jakab2018unsupervised} & 0.5 & 52.2 & 32.4 & 26.1 & 3.7 & 24.6 & 23.2  \\
  \midrule
  Ours & 81.1 & 89.8 & 39.7 & 94.2 & 93.6 & 64.4 & 77.1  \\
\midrule
    \multicolumn{8}{c}{\textbf{PCK}($\%$)}\\
\midrule
 \textbf{Method}  & Head & Shldrs & Elbws & Waist & Knees & Legs & Avg\\  
\midrule
  Zhang~\cite{Zhang2018UnsupervisedDO}  & 11.1 & 34.8 & 44.6 & 20.9 & 69.3 & 0.50 & 30.2 \\
  Jakab~\cite{jakab2018unsupervised} & 0.20 & 39.8 & 19.3 & 15.2 & 2.15 & 14.1 & 15.1  \\
  \midrule
  Ours & 51.7 & 86.3 & 43.0 & 92.2 & 83.9 & 63.5 & 70.1  \\
  \midrule
      \multicolumn{8}{c}{\textbf{Average Precision and Recall} (over OKS)}\\
\midrule
   \textbf{Method} & & $AP$ & $AP_{0.5}$  & $AP_{0.4}$ & $AR$ & $AR_{0.5}$  & $AR_{0.4}$\\ 
  \midrule  
  Zhang~\cite{Zhang2018UnsupervisedDO} & & 0.02 & 0.17 & 0.67 & 0.06 &  0.41    & 0.82  \\  
    Jakab~\cite{jakab2018unsupervised} & & 0.0 & 0.0  & 0.07 & 0.0 &  0.01 &  0.25\\
  \midrule
  Ours & & 0.22 & 0.84  & 0.95 & 0.30 &  0.91 &   0.97  \\
  \bottomrule
  \end{tabular}
\caption{Evaluation of raw discovered landmarks that correspond maximally to each ground-truth point. We report accuracy as $\%$-age of points within $d=6px$ from the ground-truth (image resolution of $128px$), the Percentage of Correct Keypoints (PCK) calculated over a threshold of $0.3$ of torso length, as well as Average Precision (AP) and Recall (AR) (commonly used to evaluate \textit{supervised} human pose estimation methods, for example \cite{Xu2022ViTPoseSV}). We also calculate AP and AR with a relaxed OKS threshold of $0.4$.}\label{humanraw}
\end{table}

We also note that due to the large degree of pose variation for human bodies, a simple linear layer does not suffice to learn a strong mapping between unsupervised and supervised landmarks. Hence, the forward error is
very high for all methods. To address this, we follow~\cite{jakab2018unsupervised} and directly evaluate the quality of raw unsupervised landmarks that are found to maximally correspond to the provided ground-truth points (calculated through the Hungarian Algorithm) for Human3.6 and PennAction databases (Table \ref{humanraw} and \ref{pennraw}). We observe that our approach can discover unsupervised landmarks that robustly track several parts of the human body (except the hands for both Human3.6M and PennAction). For Human3.6 (Table \ref{humanraw}) we can see that our method surpasses state-of-the-art methods in all reported metrics. For the challenging PennAction database that includes large pose variation and complicated backgrounds, we demonstrate higher accuracy (Table~\ref{pennraw}), whereas ~\cite{jakab2018unsupervised} completely underperforms in this setting. Note that we do not use temporal supervision to train examined methods.

\section{Conclusion} 

We presented a novel path for unsupervised discovery of object landmarks based on two ideas, namely self-training and recovering correspondence. The former helps our system improve by using its own predictions and constitutes a natural fit for training an object landmark detector starting from generic, noisy keypoints.
The latter, although being a key property of object landmarks detectors, has not been previously used for unsupervised object landmark discovery. Compared to previous works, our approach can learn view-based landmarks that are more flexible in terms of changes in 3D viewpoint, providing superior results on a variety of challenging facial and human pose datasets.

\ifCLASSOPTIONcompsoc
  \section*{Acknowledgments}
\else
  \section*{Acknowledgment}
\fi

Dimitrios Mallis’ PhD studentship is funded by The Douglas Bomford Trust.

\ifCLASSOPTIONcaptionsoff
  \newpage
\fi

    \bibliographystyle{abbrv}

\bibliography{references}

\end{document}